\documentclass[A4]{IEEEtran}
\IEEEoverridecommandlockouts

\usepackage[caption=false,font=footnotesize]{subfig} 
\usepackage{cite}
\usepackage{amsmath,amssymb,amsfonts} 
\usepackage{breqn}
\usepackage{booktabs}
\usepackage{cases}
\usepackage{graphicx}
\usepackage{multirow}
\usepackage{mathrsfs}
\usepackage{textcomp}
\usepackage{xcolor}
\usepackage{bm}
\usepackage{tikz}
\usepackage{pdfpages}
\usepackage{marvosym}
\usepackage{algorithm}
\usepackage{algorithmicx}
\usepackage{algpseudocode}

\usepackage{hyperref}
\usepackage{cleveref}

\def\BibTeX{{\rm B\kern-.05em{\sc i\kern-.025em b}\kern-.08em
    T\kern-.1667em\lower.7ex\hbox{E}\kern-.125emX}}

\begin{document}


\title{Training-Free Diffusion-Driven Modeling of Pareto Set Evolution for Dynamic Multiobjective Optimization} 
 
\author{
    Jian Guan,~\IEEEmembership{}
    Huolong Wu,~\IEEEmembership{}
    Zhenzhong~Wang,~\IEEEmembership{Member,~IEEE,}
    Gary G. Yen,~\IEEEmembership{Fellow,~IEEE,}
    and Min Jiang\textsuperscript{\textasteriskcentered},~\IEEEmembership{Senior~Member,~IEEE}
 \IEEEcompsocitemizethanks{\IEEEcompsocthanksitem
Jian Guan is with the Institute of Artificial Intelligence, Xiamen University, Xiamen, 361005, China (e-mail: guanjian@stu.xmu.edu.cn).
 
Huolong Wu, Zhenzhong Wang and Min Jiang are with the Department of Artificial Intelligence, School of Informatics, Xiamen University; the Key Laboratory of Multimedia Trusted Perception and Efficient Computing, Ministry of Education; and the Key Laboratory of Digital Protection and Intelligent Processing of Intangible Cultural Heritage of Fujian and Taiwan, Ministry of Culture and Tourism, Xiamen, 361005, China (e-mail: 31520241154526@stu.xmu.edu.cn; zhenzhongwang@xmu.edu.cn; minjiang@xmu.edu.cn).
 
Gary G. Yen is with the Department of Artificial Intelligence, Sichuan University, Chengdu, 610065, China (e-mail: gyen@okstate.edu).
	
\textit{\textsuperscript{\textasteriskcentered}Corresponding author: Min Jiang.}

	}	
}

\newcommand{\zz}[1]{{\textcolor{blue}{[zhenzhong says:] #1}}}

\bibliographystyle{IEEEtran}
\maketitle

\begin{abstract}


Dynamic multiobjective optimization problems (DMOPs) feature time-varying objectives, which cause the Pareto optimal solution (POS) set to drift over time and make it difficult to maintain both convergence and diversity under limited response time. Many existing prediction-based dynamic multiobjective evolutionary algorithms (DMOEAs) either depend on learned models with nontrivial training cost or employ one-step population mapping, which may overlook the gradual nature of POS evolution.
This paper proposes DD-DMOEA, a training-free diffusion-based dynamic response mechanism for DMOPs. The key idea is to treat the POS obtained in the previous environment as a “noisy” sample set and to guide its evolution toward the current POS through an analytically constructed multi-step denoising process. A knee-point-based auxiliary strategy is used to specify the target region in the new environment, and an explicit probability-density formulation is derived to compute the denoising update without neural training. To reduce the risk of misleading guidance caused by knee-point prediction errors, an uncertainty-aware scheme adaptively adjusts the guidance strength according to the historical prediction deviation.
Experiments on the CEC2018 dynamic multiobjective benchmarks show that DD-DMOEA achieves competitive or better convergence–diversity performance and provides faster dynamic response than several state-of-the-art DMOEAs.

\end{abstract}


\section{INTRODUCTION}
\label{INTRODUCTION}
\IEEEPARstart{D}{ynamic} multiobjective optimization problems (DMOPs) are commonly encountered in various real-world engineering applications, such as dynamic path planning \cite{RN802}, resource allocation and management \cite{Gong2020}, and industrial manufacturing \cite{KONG2013}. Their core characteristic is that the objective functions or constraints change over time. This inherent dynamicity poses a severe challenge to static multiobjective evolutionary algorithms (MOEAs) \cite{10102381, RN767, 9321762}, making them difficult to quickly and accurately track the changing true Pareto optimal front (POF) or Pareto optimal solution set (POS). 

Over the past decade, a plethora of dynamic multiobjective evolutionary algorithms (DMOEAs) have been proposed. These approaches can be primarily categorized into three types: diversity-based methods, memory-based methods, and prediction-based methods \cite{RN763}. Diversity-based methods introduce new solutions to maintain population diversity, which helps prevent premature convergence to local optima when the environment shifts \cite{RN691, RN732, RN773}. However, these methods often struggle to effectively balance maintaining diversity with achieving convergence.
Memory-based methods enable evolutionary algorithms to record useful information from past environments, guiding future search directions \cite{RN833, RN888}. Yet, a key limitation is their difficulty in handling aperiodic changes in dynamic environments. Inspired by machine learning's capabilities, prediction-based methods construct models to predict solutions for the changing environments \cite{RN690}. These include both discriminative predictors \cite{RN638}, which filter historical solutions, and generative predictors \cite{RN724, RN760}, which create new ones. Both types aim to produce high-quality initial populations by learning the dynamics of the historical POS.

\begin{figure}[h]
    \centering
    \includegraphics[width=1\linewidth]{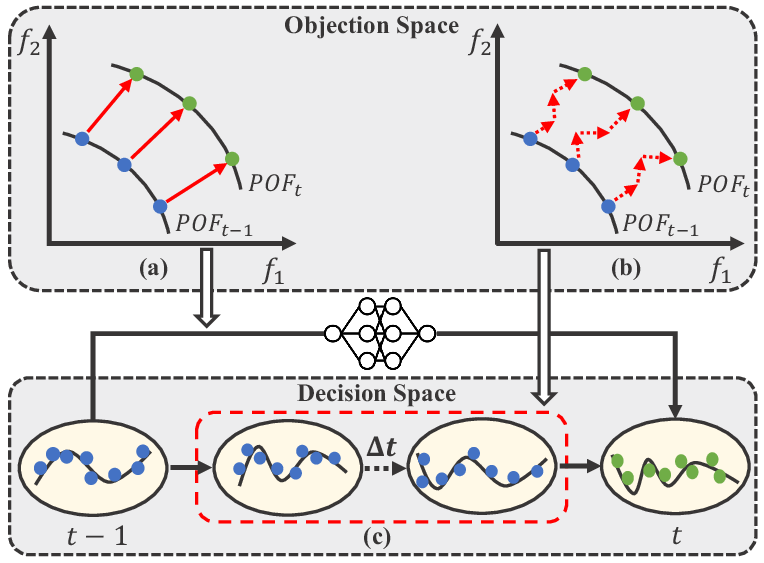}
    \caption{\textbf{Two evolutionary modes of the POS in dynamic environments}. \textbf{(a)} The single-step, direct mapping employed by existing prediction methods. \textbf{(b)} The gradual and smooth evolutionary process of the POF in the real world. \textbf{(c)} Correspondingly, in the decision space, the evolution of the POS exhibits a continuous manifold pattern. The red dashed box ($\Delta t$) indicates the key intermediate transitional states often ignored by the methods in (a).}
    \label{fig:model-1}
\end{figure}

Despite significant advancements, existing methods often generate the entire solution distribution in a single step, failing to capture the inherent continuity and smooth evolution of the population. Specifically, these methods rely on training a model to directly map the state at time $t$-1 to the final state at time $t$ as shown in Fig. \ref{fig:model-1} (a). Once trained, they predict the entire population in one step, rather than modeling the gradual, iterative nature of population evolution \cite{RN760, RN887}. However, the evolution of the population is typically a gradual process, not a single-step transition \cite{8085173, 10844572}, as illustrated in Fig. \ref{fig:model-1} (b) and (c). By embracing a multi-step, gradual approach, we can better capture the continuous, smooth trends of population evolution, allowing for more accurate predictions that reflect the true dynamics of the problem. This approach leads to a more natural alignment with evolutionary individuals, improving the prediction model’s ability to adapt to environmental changes.

Another challenge lies in the fact that most existing prediction models for DMOPs rely heavily on complex, non-lightweight machine learning techniques, which conflict with the need for rapid predictions in dynamic environments. Their response times often consume a significant amount of computational resources and time. For example, as shown in Fig. \ref{fig: rt-1-nt10-tauT10}, DM-DMOEA \cite{RN858}, which uses diffusion models, requires a total response time of up to 410.6 seconds, while DIP-DMOEA \cite{RN689}, which incorporates fully connected layers, has a response time of approximately 100 seconds. Given these limitations, there is an urgent need for prediction models that can more accurately simulate the continuous evolutionary trends of the population, while still offering fast and efficient predictions.

The inherent continuity of the population evolution suggests a process that unfolds gradually over time, rather than through abrupt transitions. This insight naturally points us toward diffusion models \cite{RN781, RN742, RN726}, which are designed to iteratively model the gradual transformation between probability distributions. These models are well-suited for our objective of capturing the continuous evolution of POS \cite{10494049}. Specifically, we view the transition from $POS_{t-1}$ to $POS_{t}$ as such a distribution transformation, where the previous $POS_{t-1}$ is viewed as noise for the new environment at $t$. In addition, unlike traditional diffusion models \cite{RN858}, which often require computationally expensive neural network training, this paper introduces a novel diffusion mechanism for solving DMOPs, named DD-DMOEA. However, while traditional diffusion models \cite{RN858} are effective in capturing gradual changes, they typically rely on computationally expensive neural network training, which hinders their suitability for real-time dynamic optimization.

In this work, we introduce a novel diffusion mechanism—DD-DMOEA—that eliminates the need for heavy training. Unlike conventional approaches, DD-DMOEA is a training-free model, leveraging explicit probabilistic computations to analytically capture the smooth transition from the ``noisy" $POS_{t-1}$ to the new $POS_{t}$. By conceptualizing the solution evolution as a denoising diffusion process, we avoid the complexity of traditional machine learning models, offering a more efficient and scalable solution. Furthermore, DD-DMOEA explicitly models intermediate transitional states of the population evolution. This allows for a more accurate representation of the continuous trends in POS evolution, enabling better tracking of the dynamic changes in the environment. The method progressively guides the distribution toward the new target regions, as estimated by knee point prediction, through explicit analytical probability calculations. The main contributions are highlighted as follows: 

\begin{enumerate}

\item We propose a novel diffusion mechanism, DD-DMOEA, which eliminates the need for heavy training typically required by traditional methods. DD-DMOEA is a training-free model that uses explicit probabilistic computations to analytically model the smooth transition from historical POS and new POS, providing an efficient solution for dynamic multiobjective optimization problems

\item Unlike existing methods that predict the entire solution distribution in a single step, DD-DMOEA adopts a multi-step, gradual approach, explicitly modeling the intermediate transitional states of population evolution. This enables the model to capture the continuous, smooth trends in POS evolution, improving accuracy and the ability to track dynamic environmental changes more effectively.

\item To alleviate the misleading risk potentially caused by knee point prediction deviations and enhance the robustness of the algorithm, we propose an uncertainty-aware diffusion guidance strategy, which significantly improves the algorithm's fault tolerance capability against complex nonlinear environmental changes.

\item Experimental studies show that DD-DMOEA provides fast and efficient predictions, significantly reducing response time by at least $\sim$51.4\% compared to selective methods, while achieving promising diversity and convergence over a wide range of state-of-the-art DMOEAs.

\end{enumerate}

The remainder of this paper is organized as follows. Section \ref{PRELIMINARIES} outlines the basic concepts of DMOPs and diffusion models, and reviews the related research work on DMOEAs. The method proposed in this paper is introduced in Section \ref{PROPOSED_ALGORITHM}. Section \ref{EXPERIMENTS} provides a comparative analysis of the performance of DD-DMOEAs through a series of experiments. Finally, Section \ref{Conclusion} summarises the conclusions of this paper and gives an outlook on future research.

\section{PRELIMINARIES} 
\label{PRELIMINARIES}
This section first describes the basic concepts and definitions of DMOPs and diffusion models, followed by a comprehensive review of the related research work on DMOEAs.

\subsection{Dynamic Multiobjective Optimization Problems}
This paper considers the minimization of DMOPs, and its mathematical definition is as follows:
\begin{equation}
\begin{aligned}
\min \boldsymbol{F}(\boldsymbol{x},t)= & (f_{1}(\boldsymbol{x},t), f_{2}(\boldsymbol{x},t), \ldots, f_{m}(\boldsymbol{x},t))^T \\
& \text { s.t. } \boldsymbol{x} \in \Omega_x, \boldsymbol{t} \in \Omega_t
\end{aligned},
\end{equation}
where $m$ is the number of objective functions. $\boldsymbol{x}=\left(x_{1}, x_{2}, \ldots, x_{n}\right)$ is an n-dimensional decision vector. $t$ is a discrete-time variable that represents environmental changes. $\Omega_x$ and $\Omega_t$ are the decision space of the variable $x$ and the time space of the variable $t$, respectively. $\boldsymbol{F}(\boldsymbol{x},t)$ represents the objective vector evaluated by $m$ objective functions at time $t$. Due to dynamic environmental changes, the POF or POS may change, so the POF and POS in dynamic environments are defined as follows: 

\textit{Definition 1 (Dynamic vector dominance)}: At time $t$, a decision vector $x_1$  is said to dominate another decision vector $x_2$, denoted $\boldsymbol{x}_1 \prec_t \boldsymbol{x}_2$, if and only if their objective functions satisfy the following conditions.
\begin{equation}
\begin{cases}
f_{i}(\boldsymbol{x}_1,t) \le f_{j}(\boldsymbol{x}_2,t), & \forall  i\in(1, \ldots, m) \\
f_{i}(\boldsymbol{x}_1,t) < f_{j}(\boldsymbol{x}_2,t), & \exists  j\in(1, \ldots, m).
\end{cases}
\end{equation}

\textit{Definition 2 (Dynamic POS (DPOS))}: At time $t$, if there does not exist any decision vector $x$ that dominates vector $x^*$, then $x^*$ is called the Pareto optimal solution, and the collection of all POS constitutes the Pareto optimal solution set.
\begin{equation}
\mathrm{DPOS}_t=\left\{\boldsymbol{x}^{*}\in \Omega_x \mid \nexists \boldsymbol{x}, \boldsymbol{x} \prec_t \boldsymbol{x}^{*}\right\} .
\end{equation}

\textit{Definition 3 (Dynamic POF (DPOF))}: The mapping of POS at time $t$ in the objective space is referred to as the Pareto optimal front.
\begin{equation}
\mathrm{DPOF}_t=\left\{\boldsymbol{F}(\boldsymbol{x}^*,t) \mid \boldsymbol{x}^*\in \mathrm{DPOS}_t \right\} .
\end{equation}

\subsection{Diffusion Model}
As a generative model based on probability distribution, the diffusion model \cite{RN781}, along with variable auto-encoder (VAE) and generative adversarial networks (GAN) \cite{GAN2014}, is the current mainstream generative artificial intelligence technology, which shows excellent performance in the field of image generation \cite{dm-image}. The model deeply investigates the intrinsic law of data evolution from simple distribution to complex distribution by simulating the change of the probability distribution of data at different time steps. From the algorithmic framework, the diffusion model mainly contains two processes: forward diffusion and reverse denoising.

In the forward diffusion process, the original data $\boldsymbol{x}_0$ gradually becomes pure Gaussian noise after multiple additions of Gaussian noise. This process can be represented as:
\begin{equation}
\boldsymbol{x}_k=\sqrt{\alpha_k}x_0+\sqrt{1-\alpha_k}\boldsymbol{\epsilon},
\label{Eq:x_t}
\end{equation}
where $\boldsymbol{x}_k$ is the data controlled by the schedule of noise $\alpha_k$ at time step $k$. $\boldsymbol{\epsilon} \sim \mathcal{N}(0,I)$ is the amount of noise at time $k$. With increasing values of $\alpha_k$, $\boldsymbol{x}_k$ gradually becomes pure Gaussian noise.

The reverse denoising process is the process of recovering the original data from noise. It mainly predicts the noise through neural networks and gradually reconstructs the original data $\boldsymbol{x}_0$ from the noisy data $\boldsymbol{x}_k$. In denoising diffusion implicit models (DDIM) \cite{RN840}, it can be expressed as:
\begin{equation}
\begin{split}
   \boldsymbol{x}_{k-1} = &\sqrt{\alpha_{k-1}} \left( \frac{\boldsymbol{x}_k - \sqrt{1 - \alpha_k} \boldsymbol{\epsilon}_\theta(\boldsymbol{x}_k, k)}{\sqrt{\alpha_k}} \right) \\
&+ \sqrt{1 - \alpha_{k-1} - \sigma_k^2} \boldsymbol{\epsilon}_\theta(\boldsymbol{x}_k, k) + \sigma_k \xi  
\end{split},
\label{Eq:x_{k-1}}
\end{equation}
where $\boldsymbol{\epsilon}_\theta(\boldsymbol{x}_k, k)$ is the noise predicted by the neural network at time step $k$. $\sigma_k$ denotes the total amount of noise at time step $k$, related to $\alpha_k$. $\xi \sim \mathcal{N}(0,I)$ is a standard Gaussian noise. 

In a nutshell, diffusion models operate through an iterative denoising process. Generating a new sample is not a direct, single-step mapping; instead, it starts from pure noise and undergoes tens to thousands of consecutive ``denoising steps" performed by the model. At each time step, the model must run a neural network (typically a U-Net architecture) to predict and remove the current step's noise. This multi-step, sequential nature inherently makes diffusion model inference slow, which directly conflicts with the need for high efficiency in solving DMOPs. Therefore, in this paper, we propose a novel training-free diffusion-inspired method to efficiently predict the POS in new environments by mimicking the core mechanism of diffusion without the typical computational overhead.

\subsection{Related Work}

Based on the DMOEAs proposed by current research\cite{RN763}, we categorize them into three main types: diversity-based, memory-based, and prediction-based methods.

\textit{\textbf{1) Diversity-based methods}}. Due to dynamic changes in the environment, previous populations may not be able to adapt to the new environmental state, resulting in loss of population diversity \cite{RN739}. To address this problem, researchers have proposed various diversity-based approaches, mainly including two categories of diversity maintenance and diversity enhancement. Earlier, Deb \textit{et al.} \cite{RN824} proposed a dynamic optimization method (DNSGA-II) based on the elitist non-dominated sorting genetic algorithm (NSGA-II) to maintain population diversity by introducing random solutions and replacing some of the mutated solutions. Similarly, Azevedo \textit{et al.} \cite{RN828} designed a diversity generator based on an immigration mechanism to maintain population diversity. In addition, the literature \cite{RN829} noted the similarity of fitness landscapes before and after environmental changes, and combined with an orthogonal operation to reduce diversity loss by using the population before environmental change directly as the initial population of the new environment. Unlike traditional methods, Peng \textit{et al.} \cite{RN825} used the knowledge and information generated by the populations before and after environmental changes to improve population diversity through guidance strategies. To predict the moving trend of the POS more accurately, Ma \textit{et al.} \cite{RN827} constructed a centroid-based difference model and generated diverse individuals within the predicted PS subregion to improve population diversity. However, due to the complexity and diversity of dynamic environments, it is difficult to accurately track true POF and POS after environmental changes by relying only on simple diversity maintenance or enhancement strategies, and thus these methods are usually used as auxiliary strategies in DMOEAs \cite{RN692}.

\textit{\textbf{2) Memory-based methods}}. With the deep research on DMOPs, scholars have found that when the new environment is similar to the historical environment, reusing the solutions in the historical environment can effectively improve the search efficiency and quality of solutions in the new environment \cite{RN729}. When a change in the environment is detected, Jiang \textit{et al.} \cite{RN831} adopt a steady-state response mechanism to reuse some of the high-quality solutions in the historical environment to relocate the POF quickly. Liang \textit{et al.} \cite{RN830} propose two different response strategies by analyzing the similarity between the current and the historical environment. If the current environment is similar to the historical environment, the stored high-quality solutions are directly reused. Otherwise, the difference in population centers in the first two consecutive environments is used to relocate individuals to the new environment. Similarly, an algorithm based on the intensity of environmental changes was proposed in the literature \cite{RN832}, which uses historical information as a reference to guide the evolution of the population towards a promising region of the decision space. Unlike previous studies, Zou \textit{et al.} \cite{RN834} emphasized the influence of the environment on evolutionary individuals and proposed a dynamic evolutionary environment model to help populations better adapt to new environments through the rational use of knowledge and information accumulated during environmental change. However, the memory-based approach is highly dependent on the similarity between the new environment and the historical environment. When the environment changes drastically, this approach may not be able to effectively respond to the challenges of the new environment \cite{RN728, RN688}.  

\textit{\textbf{3) Prediction-based methods}}. As the current mainstream research direction, prediction-based methods mainly analyze the potential patterns of historical environment changes to generate high-quality solutions for new environments \cite{RN815, RN835}. Jiang \textit{et al.} \cite{RN544} firstly pointed out that the POF or POS at different times may not be the same, but there exists a correlation and conforms to the characteristic of non-independent identical distribution. They utilize transfer learning as a predictor to accelerate the evolutionary process by reusing historical experience, thus generating an effective initial population. To solve the negative transfer problem, Jiang \textit{et al.} \cite{RN547} further proposed a pre-search strategy to avoid the negative transfer phenomenon due to individual aggregation by screening high-quality individuals. In a recent study \cite{RN690}, Lin \textit{et al.} classified predictors into generative predictors and discriminative predictors, and proposed a DMOEA with a knowledge transfer and maintenance mechanism, which effectively alleviates the negative transfer problem and improves the optimization efficiency. For generative predictors, Feng \textit{et al.} \cite{RN726} learn the changes of two continuous dynamic environments through a linear denoising autoencoder (DAE), which reveals the mapping relationship between the non-dominated solutions, and thus tracks the moving direction of the optimal solution. In contrast, Hu \textit{et al.} \cite{RN727} proposed a Mahalanobis distance-based approach for DMOPs with random changes and used linear and nonlinear autoencoders to predict a new solution set to accelerate population convergence and maintain diversity. In addition, Li \textit{et al.} \cite{RN760} combined a GAN with autoencoders to generate high-quality initial populations for a new environment by training generative models using POS in the current environment and transferring them to the new environment. Although the predictor-based method can generate high-quality solutions for the new environment, it depends on the training process of the model, making it difficult for the algorithm to respond to the dynamic changes of the environment immediately, which is still an important challenge in the current research.

\begin{figure*}[h] 
  \centering   
  \includegraphics[width=0.9\linewidth]{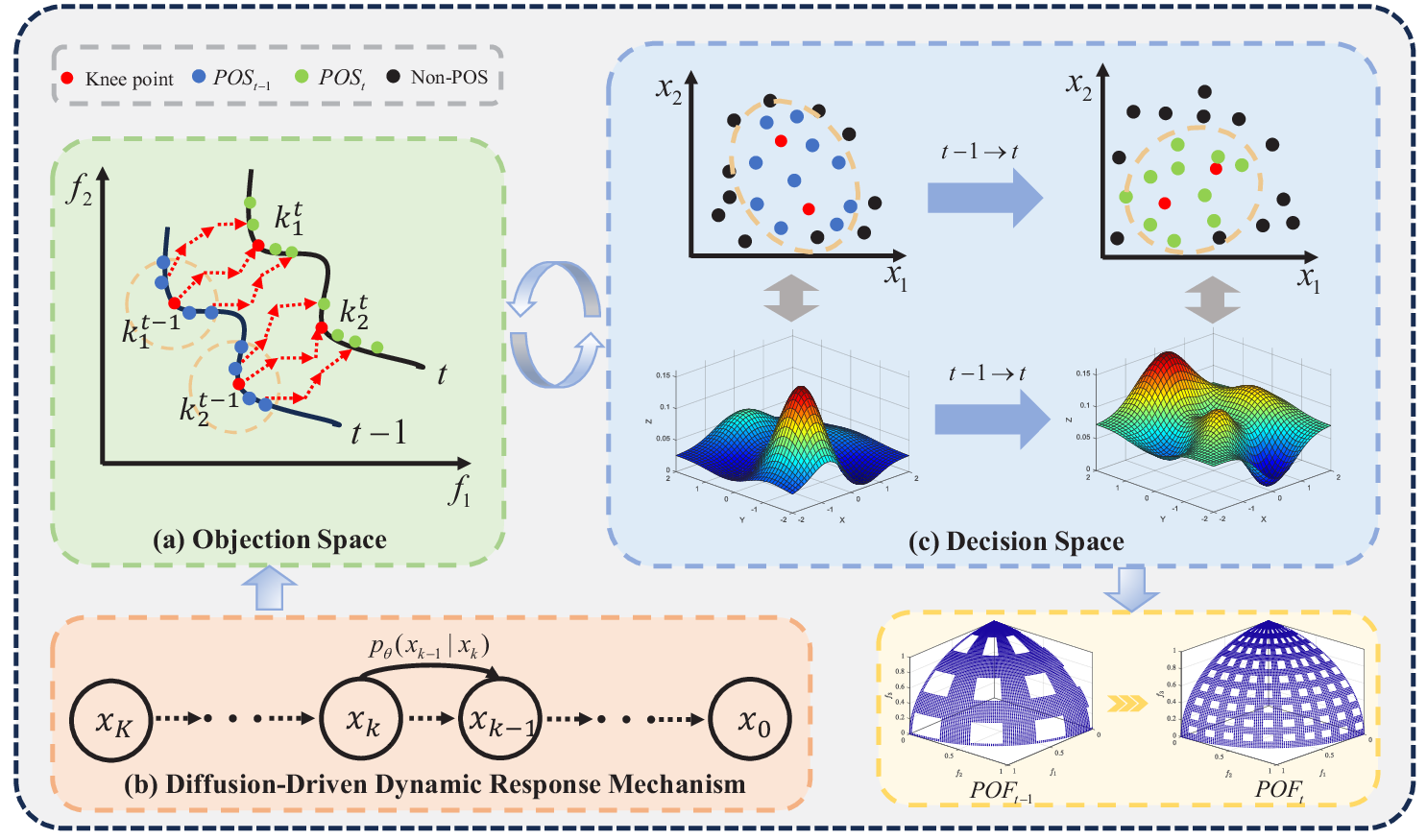}
  \caption{\textbf{Overview of the proposed DD-DMOEA.} \textbf{(a)} The AKP strategy partitions the objective space and predicts the positions of the knee points ($k^t_i$) at the moment $t$. This prediction assists in generating an approximate target POS distribution based on $POS_{t-1}$. \textbf{(b)} The DDM treats the $POS_{t-1}$ as a noisy distribution, and a training-free diffusion model executes an efficient and progressive denoising process. \textbf{(c)} As the environment changes ($t$-1 $\to$ $t$), this denoising process gradually transforms the ``noisy" old population into a refined one adapted to the new environment, enabling rapid tracking of the POF's evolution ($POF_{t-1} \to POF_{t}$).}
  \label{fig: Framework figure}
\end{figure*}

\section{PROPOSED METHOD} 
\label{PROPOSED_ALGORITHM}

\subsection{Overall Framework}
\label{Overall Framework}

In our work, we propose a novel training-free diffusion method to generate POS for new environments. A key challenge is that, unlike traditional diffusion models, which require knowledge of the true solution distribution for their generation process, the true POS distribution in a new dynamic environment is unknown. To address this, we first introduce an auxiliary strategy based on knee point estimation (AKP) to generate an approximate POS distribution, which serves as the true distribution for the diffusion process. Subsequently, a diffusion-driven dynamic response mechanism (DDM) takes the noisy POS distribution from the previous time step. Through an efficient and progressive denoising process, the DDM generates high-quality solutions adapted to the new environment. The overall framework of DD-DMOEA is illustrated in Fig.~\ref{fig: Framework figure}.


\begin{algorithm}[h]
    \caption{DD-DMOEA}
    \label{alg: DD-DMOEA}
    \begin{algorithmic}[1] 
    \renewcommand{\algorithmicrequire}{\textbf{Input:}}
    \renewcommand{\algorithmicensure}{\textbf{Output:}}
        \Require \parbox[t]{0.93\linewidth} {\strut The dynamic problem $F(\boldsymbol{x},t)$, population size $N$, the number of subspaces $N_s$. \strut} 
        \Ensure \parbox[t]{0.9\linewidth} {\strut An approximate POS and POF in different environments.\strut}
        \State Initialize related parameters and set $t$=1;      
        \While{\textit{the environment has changed}}
                \If{\textit{$t$ $\leq$ 2}}
                    \State Randomly initialize a population \textit{initPop};
                    \State $POS_t$=\textbf{SMOEA}(\textit{initPop}, $F(\boldsymbol{x},t)$);
                \Else
                    \State $knees$=\textbf{AKP}($knee_{t-1}$, $knee_{t-2}$, $F(\boldsymbol{x},t)$);
                    \State $Pop_{pred}$=\textbf{DDM}($POS_{t-1}$, $knees$, $N_s$, $\psi^t$);
                    \State \textit{initPop}=$Pop_{pred}$ $\cup$ $Pop_{last}$ $\cup$ $Pop_{rand}$;
                    \State $POS_t$=\textbf{SMOEA}(\textit{initPop}, $F(\boldsymbol{x},t)$);
                \EndIf
                \State \parbox[t]{0.93\linewidth} {\strut Evaluate the true knee points at the moment $t$ by $POS_t$;\strut}
                \State $t=t+1$;
                \State \parbox[t]{0.93\linewidth} {\strut Calculate the adaptive variance $\psi^t$ of the current Gaussian distribution;\strut}
           
        \EndWhile         
    \end{algorithmic}
\end{algorithm}

The pseudocode of the proposed DD-DMOEA is presented in Algorithm \ref{alg: DD-DMOEA}. Initially, relevant parameters and the discrete-time variable $t$=1 are set. Subsequently, the algorithm enters the main loop at line 2, which corresponds to the dynamic response phase. In this process, if the discrete-time variable $t \leq 2$, the algorithm randomly initializes a population $initPop$ for the current environment in line 4, and obtains an approximate POS ($POS_t$) at moment $t$ by the static MOEA (SMOEA) in line 5. Note that the SMOEA used in this paper is MOEA/D. Instead, the algorithm will execute lines 7-10, which is also the core part of the algorithm in this paper. First, the AKP strategy in line 7 predicts the knee points information $knees$ at the current moment $t$ based on the knee points ($knee_{t-1}$ and $knee_{t-2}$) of the previous two moments. Then, in line 8, the DDM considers the $POS_{t-1}$ at moment $t$-1 to be a noisy distribution, through an efficient and progressive denoising process, and generates high-quality solutions ($Pop_{pred}$) adapted to the environment at moment $t$. To maintain population diversity, the initial population at the current moment, $initPop$, consists of three parts, which are $Pop_{pred}$ predicted by the DDM, the POS at the previous moment ($Pop_{last}$), and the randomly generated population $Pop_{rand}$. Finally, the approximate POS at the current moment ($POS_t$) is obtained by SMOEA in line 10. Since the knee points information of AKP is obtained by prediction, the true knee points information of moment $t$ is evaluated to satisfy the iterative requirement in line 12. Given that the guidance of DDM is highly dependent on the accuracy of the predicted knee points, this paper adopts an uncertainty-aware diffusion guidance strategy at line 14, thereby enhancing the robustness of the algorithm. The details of the AKP and DDM methods are described in Sections \ref{AKP} and \ref{DDM}, respectively. 

\begin{figure}[h] 
  \centering   
  \includegraphics[width=0.8\linewidth]{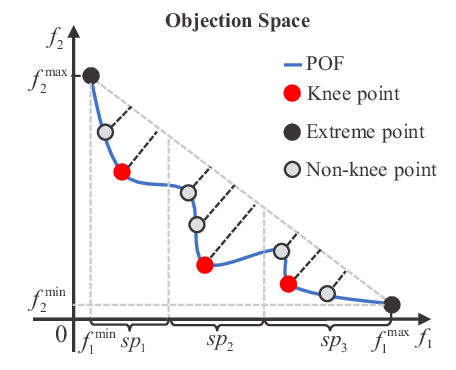}
  \caption{Illustration of dividing objection space by the knee point. The number of subspaces here is 3.}
  \label{fig:kneePoints}
\end{figure}

\subsection{Auxiliary Strategy Based on Knee Point Estimation}
\label{AKP}

The original diffusion mechanism requires knowledge of the true POS distribution to guide its generation process. However, this true POS distribution for the future environment is inherently unknown. To overcome this limitation, we first introduce an AKP strategy, detailed in Algorithm \ref{alg: AKP}. The core of the AKP strategy is to approximate the true POS distribution by estimating its knee points, which are special points on the POF where the trade-off between objective functions changes significantly. This approximation provides the necessary reference for the DDM \cite{RN699, RN545}.

Specifically, to identify and track the evolution of the entire POF, we first partition the objective space into multiple subspaces $sp_i$ and determine a knee point within each, as shown in Fig.~\ref{fig:kneePoints}, and the size of each subspace is decided as follows. Furthermore, by focusing only on a few high-quality knee points, the waste of computational resources is significantly reduced. 

\begin{equation}
S^i_{m,t}=\frac{f^{max}_{m,t} - f^{min}_{m,t}}{N_s}.
\end{equation}

The upper and lower boundaries of each subspace are $\text{UB}^i_{m,t}=f^{min}_{m,t}+i \cdot S_{m,t}$ and $\text{LB}^i_{m,t}=f^{min}_{m,t}+(i-1) \cdot S_{m,t}$, respectively.

\begin{algorithm}[h]
    \caption{Auxiliary Strategy Based on Knee Point Estimation (AKP)}
    \label{alg: AKP}
    \begin{algorithmic}[1] 
    \renewcommand{\algorithmicrequire}{\textbf{Input:}}
    \renewcommand{\algorithmicensure}{\textbf{Output:}}
         \Require \parbox[t]{0.93\linewidth} {\strut The dynamic problem $F(\boldsymbol{x},t)$, population size $N$, the POS $POS_{t-1}$ and $POS_{t-2}$ at moments $t$-1 and $t$-2, the number of subspaces $N_s$. \strut}
        \Ensure \parbox[t]{0.93\linewidth} {\strut Knee points in the next environments $knees$. \strut}
        \State Randomly initialize a population \textit{initPop};  
        \State Obtain the $knee_{t-1}$, $knee_{t-2}$ by $POS_{t-1}$, $POS_{t-2}$ respectively;
        \For{$i$=1 \textit{to} $N_s$}
            \For {$j$=1 \textit{to} $n$-1}
                \State Obtain the direction vectors $v_i$ between the knee  
                \Statex \qquad \quad points through Eq.(\ref{Eq: direction vector});
                \State Calculate the $j$-th angular coordinates $\beta^j_i$ of $v_i$ 
                \Statex \qquad \quad through Eq.(\ref{Eq: angular coordinate});
                \State Calculate the deflection angles $\theta^j_i$ via Eq.(\ref{Eq.theta});
            \EndFor
            \State \parbox[t]{0.93\linewidth} {\strut The knee points $k^i_t$ at moment $t$ and $u_i$ are obtained through Eq.(\ref{Eq: kt}) and (\ref{Eq:u}), respectively;\strut}
        \EndFor
        \State \textbf{return} $knees$; 
    \end{algorithmic}
\end{algorithm}

Subsequently, using the knee points from the two preceding time steps ($knee_{t-1}$ and $knee_{t-2}$), the algorithm calculates the motion direction vector $v_i$, for the knee point in the $i$-th subspace. Ultimately, a trend prediction model (TPM) \cite{RN545} is constructed by tracking these motion trajectories across consecutive environments.
\begin{equation}
v_i=knee_{t-1}^i - knee_{t-2}^i.
\label{Eq: direction vector}
\end{equation}

To precisely predict the moving direction of the knee point, the direction vector is denoted as $v_i=\left\{ \beta_i^j,r \mid j \in [1, n-1] \right\}$ in polar coordinates, where $\beta^j_i$ and $r$ denote the $j$-th angular and radial coordinates of the direction vector $v_i$, respectively. Meanwhile, to facilitate the calculation, the TPM sets the radial coordinate $r$ to be the same as $|v_i|$. For the $j$-th angular coordinate of the $i$-th direction vector $v_i$ is calculated as follows:
\begin{equation}
\beta^j_i=\arctan(\frac{\sqrt{\sum_{d=j+1}^{n}(v^d_i)^2}}{v^j_i}),
\label{Eq: angular coordinate}
\end{equation}
where $v^d_i$ represents $d$-th component vector of $i$-th direction vector $v_i$. $n$ is the size of the decision vector. 

When the environment changes, different angular coordinates produce multiple deflection angles $\theta^j_i$. Here, the deflection angle with the highest probability is selected by the probability density function in the following formula.
\begin{equation}
\theta_i^j \sim \text{TPM}(v_i) = \frac{e^{-\text{sign}(\theta_i^j) \cdot \frac{\theta_i^j}{|v_i|}}}{\int_{-\pi}^{0} e^{\frac{\theta_i^j}{|v_i|}} d\theta_i^j + \int_{0}^{\pi} e^{-\frac{\theta_i^j}{|v_i|}} d\theta_i^j},
\label{Eq.theta}
\end{equation}
where $\theta_i^j$ denotes the $j$-th deflection angle of the $i$-th direction vector $v_i$. According to the characteristics of DMOP, we argue that the moving direction of the knee point will move in a certain direction with high probability; that is, when the deflection angle is $0^\circ$, it is the most likely moving direction of the knee point. Meanwhile, $0^\circ$ has a high probability density value in Eq. (\ref{Eq.theta}).

Finally, based on the above deflection angle $\theta_i^j$ and the knee point $knee^{t-1}$ at moment $t$-1, we can calculate the knee point in the $i$-th subspace at moment $t$ as follows.
\begin{equation}
knee^i_t=knee^i_{t-1}+r\times  \Theta_i.
\label{Eq: kt}
\end{equation}

The vector $\Theta^j_i$ represents the $j$-th component vector of the vector $\Theta_i$, which is obtained by the following formula.
\begin{equation}
\Theta^j_i=
\begin{cases}
\cos(\beta^j_i+\theta_i^j), & j=1 \\
\prod_{d=1}^{j-2} \sin(\beta_i^d + \theta_i^d) \cos(\beta_i^{j-1} + \theta_i^{j-1}), & 1 < j < n \\
\prod_{d=1}^{j-1} \sin(\beta_i^d + \theta_i^d), & j = n
\end{cases}.
\label{Eq:u}
\end{equation}

\subsection{Diffusion-Driven Dynamic Response Mechanism}
\label{DDM}
The generation process of traditional diffusion models relies on an iterative denoising sequence, often requiring hundreds to thousands of steps, with each step invoking a deep neural network for inference \cite{RN742}. This inherent complexity and computational overhead render them too slow for DMOPs. To leverage the advantages of diffusion models while avoiding this overhead, this paper proposes a training-free DDM. Specifically, we consider the $POS_{t-1}$ from the previous time step as a noisy version of the true POS in the current new environment. Therefore, each denoising step directly corresponds to a temporal state in the DMOP's evolution process. The DDM efficiently and accurately tracks the dynamically changing optima by progressively transforming these noisy solutions into solutions adapted to the new environment.


\begin{algorithm}[h]
    \caption{Diffusion-Driven Dynamic Response Mechanism (DDM)}
    \label{alg: DDM}
    \begin{algorithmic}[1] 
    \renewcommand{\algorithmicrequire}{\textbf{Input:}}
    \renewcommand{\algorithmicensure}{\textbf{Output:}}
        \Require \parbox[t]{0.93\linewidth} {\strut The dynamic problem $F(\boldsymbol{x},t)$, the $POS_{t-1}$ at moment $t$-1, the number of subspaces $N_s$, the size of the decision vector $n$, total time step $K$, density function $f(\cdot)$, dynamic variance $\psi^t$. \strut}
        \Ensure  \parbox[t]{0.93\linewidth} {\strut The predicted population $Pop_{pred}$ at moment $t$.\strut}
        \State $\alpha_k = \{\alpha_k \mid \alpha_k = \frac{1}{2}[\text{cos}(\frac{k\pi}{K})+1], k \in [0, K]\}$;      
        \For{$i$=1 \textit{to} $N_s$}     
            \State $[\boldsymbol{x}^{i,1}_k,\boldsymbol{x}^{i,2}_k, \dots ,\boldsymbol{x}^{i,n}_k ] \gets POS^i_{t-1}$;
            \For{$k$=$K$ \textit{to} 2}  \Comment{$K \gg 2$}
                \State $\sigma^i_k = \sqrt{\frac{1-\alpha^i_{k-1}}{1-\alpha^i_k}\times(1-\frac{\alpha^i_k}{\alpha^i_{k-1}})}$;
                \State $\Phi=\boldsymbol{x}^{i,j}_k$; \Comment{$j \in [1,n]$}
                \State $\mathcal{Z} = \sum\limits_{j=1}^{n} f(\Phi,\psi_i^t) \mathcal{N}(\Phi; \sqrt{\alpha_k} \Phi, 1 - \alpha_k)$; 
                \State $\hat{\boldsymbol{x}}_0 = \frac{1}{\mathcal{Z}} \sum\limits_{j=1}^{n} f(\Phi, \psi_i^t) \mathcal{N}(\Phi; \sqrt{\alpha_k} \Phi, 1 - \alpha_k) \Phi$;
                \State $\boldsymbol{x}^{i,j}_{k-1} = \sqrt{\alpha_{k-1}} \tilde{\boldsymbol{x}}_0 + \sqrt{1 - \alpha_{k-1} - \sigma_k^2} \tilde{\boldsymbol{\epsilon}} + \sigma_k \xi$;
            \EndFor
            \State $Pop^i_{pred} \gets [\boldsymbol{x}^{i,1}_1,\boldsymbol{x}^{i,2}_1, \dots ,\boldsymbol{x}^{i,n}_1 ]$;
        \EndFor
        \State \textbf{return} $Pop_{pred}$;
    \end{algorithmic}
\end{algorithm}

The core of this training-free mechanism is detailed in Algorithm \ref{alg: DDM}. DDM completely bypasses the complex, neural network-based training required by traditional diffusion models. Instead, it analytically implements the reverse denoising process using probability density calculations. In this process, we consider the historically acquired $POS_{t-1}$ as the initial ``noisy", low-quality distribution. Meanwhile, the knee points estimated by the AKP strategy define the approximate target region ($POS_t$). The DDM's denoising process must then analytically guide the evolution of this ``noisy" distribution toward this high-quality target region. Before deriving the analytical process, we first rewrite the basic equations of the standard diffusion model. From the forward process (Eq. \ref{Eq:x_t}), we can derive the expressions for $x_0$ and noise $\epsilon$:
\begin{equation}
    \boldsymbol{x}_0=\frac{\boldsymbol{x}_k-\sqrt{1-\alpha_k}\boldsymbol{\epsilon}}{\sqrt{\alpha_k}}, \quad \boldsymbol{\epsilon}=\frac{\boldsymbol{x}_k-\sqrt{\alpha_k}\boldsymbol{x}_0}{\sqrt{1-\alpha_k}}
\label{Eq: x_0}.
\end{equation}

Since the noise in the reverse denoising process is obtained by a neural network prediction. Therefore, we can substitute $\tilde{\boldsymbol{\epsilon}}=\boldsymbol{\epsilon}_\theta(\boldsymbol{x}_k, k)$ into Eq. (\ref{Eq:x_{k-1}}), and simplify Eq. (\ref{Eq:x_{k-1}}) to the following format.
\begin{equation}
\boldsymbol{x}_{k-1} = \sqrt{\alpha_{k-1}} \tilde{\boldsymbol{x}}_0 + \sqrt{1 - \alpha_{k-1} - \sigma_k^2} \tilde{\boldsymbol{\epsilon}} + \sigma_k \xi.
\label{Eq:x_{k-1}_2}
\end{equation}


To derive this analytical process, we revisit the essence of the reverse denoising process, which is the probability density estimation of $x_0$ given $x_k$. This estimation can be computed via Bayes' theorem.
\begin{equation}
\begin{split}
    p(\boldsymbol{x}_0 = \boldsymbol{x} | \boldsymbol{x}_k) &= \frac{p(\boldsymbol{x}_k | \boldsymbol{x}_0 = \boldsymbol{x}) p(\boldsymbol{x}_0 = \boldsymbol{x})}{p(\boldsymbol{x}_k)} \\ &= \frac{p(\boldsymbol{x}_k | \boldsymbol{x}) f(\boldsymbol{x})}{p(\boldsymbol{x}_k)},
\label{Eq: Bayes theorem}
\end{split}
\end{equation}

where $p(\boldsymbol{x}_0 = \boldsymbol{x})$ is the forward diffusion process, following a Gaussian distribution $\mathcal{N}(x_{k};\sqrt{\alpha_{k}}x,1-\alpha_{k})$. $f(x) \equiv p(x_0=x)$ is the prior probability of $x_0$. This is the key to our ``training-free" guidance. In the DMOP context, $f(x)$ represents the desired probability density of a solution $\boldsymbol{x}$ belonging to the new $POS_t$. We model this as a Gaussian distribution centered on the knee points ($\mu$) predicted by the AKP strategy, and the standard deviation $\psi$ is specifically set in subsection \ref{Adaptive variance}. Solutions with a higher $f(x)$ value are closer to the high-quality target region of the new environment. $p(x_k)$ is a normalization term.


Traditional diffusion models use a complex neural network $\epsilon_{\theta}(\boldsymbol{x}_{k},k)$ to learn the prediction of the noise $\tilde{\epsilon}$ at each step of the reverse process. Our goal is to bypass this training process by analytically calculating the expected value of $x_0$, and denoted as $\hat{\boldsymbol{x}}_0$. It can be obtained by a weighted average over all possible $\boldsymbol{x}$.
\begin{equation}
\begin{split}
   \hat{\boldsymbol{x}}_0(\boldsymbol{x}_k, \alpha, k) &= \sum_{\boldsymbol{x} \in s(\boldsymbol{x})} p(\boldsymbol{x}_0 = \boldsymbol{x} | \boldsymbol{x}_k) \boldsymbol{x} \\ &= \sum_{\boldsymbol{x} \in s(\boldsymbol{x})} f(\boldsymbol{x}) \frac{\mathcal{N}(\boldsymbol{x}_k; \sqrt{\alpha_k} \boldsymbol{x}, 1 - \alpha_k)}{p(\boldsymbol{x}_k)} \boldsymbol{x},
\label{Eq: target x0} 
\end{split}
\end{equation}

In our DDM mechanism, this sampling set $s(\boldsymbol{x})$ is precisely the historical population from the previous time step, $POS_{t-1}^{i}$. By substituting Eq. (\ref{Eq: Bayes theorem}) into Eq. (\ref{Eq: target x0}) and replacing $p(\boldsymbol{x}_k)$ with the normalization constant $\mathcal{Z}$, we get the analytical estimate of $x_0$:
\begin{equation}
\hat{\boldsymbol{x}}_0(\boldsymbol{x}_k, \alpha, k) = \frac{1}{\mathcal{Z}} \sum_{\boldsymbol{x} \in \boldsymbol{x}_k} f(\boldsymbol{x}) \mathcal{N}(\boldsymbol{x}_k; \sqrt{\alpha_k} \boldsymbol{x}, 1 - \alpha_k) \boldsymbol{x},
\label{Eq:17}
\end{equation}
where the normalization term $\mathcal{Z}$ can be written in the following format.
\begin{equation}
\mathcal{Z} = \sum_{\boldsymbol{x} \in \boldsymbol{x}_k} f(\boldsymbol{x}) \mathcal{N}(\boldsymbol{x}_k; \sqrt{\alpha_k} \boldsymbol{x}, 1 - \alpha_k).
\label{Eq:Z}
\end{equation}

By Eq. (\ref{Eq:17}), we consider that the estimate of the new $POS_t$ (i.e., $\hat{\boldsymbol{x}}_0$) is a weighted average of all solutions in $POS_{t-1}$ (i.e., $\boldsymbol{x}_k$). The weight $f(\boldsymbol{x})$ is determined by their probability density of being near the new environment's knee points. This achieves the goal of using the AKP-predicted target region to guide the evolutionary direction of $POS_{t-1}$. With the analytical estimate $\hat{\boldsymbol{x}}_0$, we can then calculate the corresponding ``training-free" noise $\tilde{\boldsymbol{\epsilon}}$ by Eq. (\ref{Eq: x_0}).
\begin{equation}
\tilde{\boldsymbol{\epsilon}}(\boldsymbol{x}_k, \alpha, k) = \frac{\boldsymbol{x}_k - \sqrt{\alpha_k} \tilde{\boldsymbol{x}}_0(\boldsymbol{x}_k, \alpha, k)}{\sqrt{1 - \alpha_k}}.
\label{Eq:epsilon}
\end{equation}

Finally, we substitute this analytically computed $\hat{\boldsymbol{x}}_0$ and $\tilde{\boldsymbol{\epsilon}}$ back into Eq. (\ref{Eq:x_{k-1}_2}). This completes one denoising step, i.e., one evolutionary state transition.

In summary, DDM is an iterative and analytical evolutionary process, detailed in Algorithm \ref{alg: DDM}. The core of this strategy is to reframe the evolution from $POS_{t-1}$ to $POS_t$ as a multi-step denoising transformation from a ``noisy" state to a clear state. The inputs to this mechanism are twofold: the $POS_{t-1}$ from the previous time step (treated as the initial noisy samples $x_K$) and the knee points predicted by the AKP strategy, which are used to define the high-quality target region of the new $POS_t$. DDM's training-free nature is realized by bypassing neural networks; instead, it performs the evolution through $K$ iterative analytical calculations. At each of the $K$ steps, DDM utilizes Eq. (\ref{Eq:17}) and Eq. (\ref{Eq:epsilon}) for explicit probability density calculations, thereby analytically estimating the expected target $\hat{\boldsymbol{x}}_0$ and the corresponding noise $\tilde{\boldsymbol{\epsilon}}$ for the current state. These analytically computed terms are then substituted into the reverse denoising formula (i.e., Eq. (\ref{Eq:x_{k-1}_2})) to execute the transformation. This formula consists of two key components: a guidance term ($\sqrt{\alpha_{k-1}}\hat{\boldsymbol{x}}_{0}+\sqrt{1-\alpha_{k-1}-\sigma_{k}^{2}}\tilde{\boldsymbol{\epsilon}}$), which uses the calculated $\hat{x}_0$ and $\tilde{\boldsymbol{\epsilon}}$ to drive the population progressively closer to the high-quality target region, and a diversity term ($\sigma_{k}\xi$), which is a random Gaussian noise to maintain essential population diversity. Finally, through $K$ iterations (that is, the intermediate transitional state ignored by the ``one-step'' methods), DDM efficiently and gradually transforms the ``noisy" historical population into a high-quality predicted population, $Pop_{pred}$, adapted to the current new environment.

\subsection{Uncertainty-Aware Diffusion Guidance Strategy}
\label{Adaptive variance}
The guidance effectiveness of DDM is highly dependent on the accuracy of these predicted knee points. In complex dynamic environments, environmental changes may exhibit nonlinear or irregular characteristics, leading to deviations between the predicted knee points and the true $POS_t$. In such cases, adopting a fixed standard deviation $\psi$ may mislead the evolutionary direction of the population. An overly small $\psi$ constructs a narrow target region, forcibly guiding the population to incorrect local optima, while an overly large $\psi$ weakens the guidance capability, resulting in slow convergence.

To address this dependency issue and enhance the robustness of the algorithm, this paper proposes an uncertainty-aware diffusion guidance strategy. Unlike traditional fixed parameter settings, this strategy utilizes the prediction error from historical time steps as a feedback signal to dynamically adjust the guidance intensity of the current diffusion process. Specifically, for the $i$-th subspace, its standard deviation $\psi_i^t$ at time $t$ is calculated based on the prediction uncertainty at the previous time step $t-1$. We first define the prediction error as the Euclidean distance between the predicted knee point $knee_{pred}^{i,t-1}$ at time $t-1$ and the true knee point $knee_{true}^{i,t-1}$ obtained after optimization. Subsequently, the adaptive variance $\psi_i^t$ is updated via the following formula:
\begin{equation}
\begin{split}
   E_i^{t-1} =\left \|knee_{pred}^{i,t-1}-knee_{true}^{i,t-1}\right \|_2 \\
   \psi_i^t =\text{min}(\psi_{max},\text{max}(\psi_{min}, \psi_{min}+\lambda \cdot E_i^{t-1})).
   \label{psi}
\end{split}
\end{equation}

Here, $\psi_{min}$ and $\psi_{max}$ are the lower and upper bounds of the standard deviation, set to 0.1 and 0.5, respectively, as detailed in subsection \ref{Parameter Sensitivity Analysis}. $\lambda$ is the sensitivity coefficient, which is set to 2 in this paper. Simply put, when the prediction is accurate, $\psi$ decreases to strengthen the guidance toward the target. Conversely, when the prediction error is high, $\psi$ increases to smooth the target distribution, thereby enhancing population diversity and reducing the risk of incorrect convergence.

Overall, this strategy effectively circumvents the misleading risk potentially caused by knee point prediction deviations, significantly enhancing the algorithm's fault tolerance capability against complex nonlinear environmental changes. At the same time, this characteristic of dynamically adjusting guidance precision ensures that DD-DMOEA can consistently maintain a high-quality population distribution while retaining its advantage of rapid response.

\subsection{Computational Complexity Analysis}
\label{Complexity Analysis}
The computational complexity of DD-DMOEA has two main components: the static optimizer MOEA/D and the dynamic response strategy. We know that the computational complexity of MOEA/D is $O(m\times N^2)$ through \cite{RN689}, where $m$ and $N$ denote the number of objectives and the population size in a multiobjective optimization problem, respectively. In the method presented in this paper, AKP and DDM are the main reasons for the computational complexity. The computational complexity of AKP and DDM is observed to be $O(n\times N_s)$ and $O(N_s\times K)$ by Algorithm \ref{alg: AKP} and Algorithm \ref{alg: DDM}, respectively, where $n$ and $N_s$ are the size of the decision vector and the number of subspaces, respectively, and $K$ is the total time step of the diffusion model. Since $K$ is much larger than $n$, the computational complexity of the dynamic response of DD-DMOEA is $O(N_s \times K)$.

\section{EXPERIMENTS} 
\label{EXPERIMENTS}
This section validates the effectiveness of the DD-DMOEA algorithm proposed in this paper by a series of experiments. First, we describe in detail the experimental setup, including the benchmark problems, comparison algorithms, parameter settings, and performance metrics. Subsequently, we compare and analyze the experimental results of different algorithms and further evaluate the runtime efficiency of each algorithm. Meanwhile, to investigate the role of each module in the algorithms and the impact of key parameters, we design ablation studies and conduct sensitivity analyses on the relevant parameters. Finally, the dynamic optimization capability of DD-DMOEA is verified on a real-world application.

\subsection{Benchmark Problems}
The benchmark problems used in this study were selected from the 14 test functions in the CEC2018 dynamic multi-objective optimization competition \cite{RN694}. It contains 9 bi-objective problems and 5 tri-objective problems, and these test functions can comprehensively reflect different types of DMOPs in the real world. Detailed descriptions of the benchmark problems can be found in Table S-I in the supplementary material. To accurately simulate the dynamic changes of real-world environments, this study adopts the discrete-time variable $t =\frac{1}{n_t}\left \lfloor \frac{\tau}{\tau_t} \right \rfloor$ to characterize environmental changes, where $n_t$ denotes the severity of the change, $\tau$ is the number of iterations, $\tau_t$ denotes the frequency of change, and $\left \lfloor \cdot \right \rfloor $ is the rounding operator. To adequately evaluate the performance of the algorithm in different dynamic environments, four groups of dynamic environments with different change characteristics are designed in this study: ($n_t=10, \tau_t=10$), ($n_t=5, \tau_t=10$), ($n_t=10, \tau_t=5$), and ($n_t=5, \tau_t=5$). Each group contains 30 environmental changes, that is, the value of $\tau$ is $30 \times \tau_t$, to verify the effectiveness of the dynamic response method.

\subsection{Comparative Algorithms and Parameter Settings}
The proposed DD-DMOEA was experimentally compared against six state-of-the-art algorithms, including KT-DMOEA \cite{RN545}, IGP-DMOEA \cite{RN686}, KTM-DMOEA \cite{RN690}, DIP-DMOEA \cite{RN689}, PSGN \cite{RN728}, VARE \cite{RN887}, and DM-DMOEA \cite{RN858}. These methods are all advanced prediction-based algorithms that utilize various machine learning or statistical models to respond to environmental changes. 
Their detailed information is introduced in the supplementary material.

To ensure the fairness of the experiment, this study uniformly adopts MOEA/D as the SMOEA framework for all compared algorithms. In terms of parameter settings, the population size is set to 100 for bi-objective optimization problems, while it is expanded to 150 for tri-objective optimization problems. The dimensionality of decision variables is uniformly set to 10. The crossover operator and the variance operator in MOEA/D are kept the same in all the compared algorithms. Specifically, in the proposed DD-DMOEA algorithm, the number of subspaces $N_s$ is set to 5, the total time steps $K$ of the diffusion model is set to 100, and the standard deviation $\sigma$ of the Gaussian probability density function ranges from [0.1, 0.5]. Considering the unknown prior information of sample point distribution, this study employs the kernel density estimation (KDE) method to calculate the probability density $P(\boldsymbol{x}_t)$ for each sample point. To obtain reliable statistical results, each comparison algorithm was run independently 20 times, and the experimental results were averaged. All experiments were done on Windows 11 with 128GB RAM and Intel Core i9-12900K CPU, and the algorithm implementation is based on MATLAB 2024b development environment.  

\subsection{Performance Metrics}
Two performance metrics are used in this experiment to comprehensively evaluate the convergence and diversity of the algorithms in DMOPs, namely MIGD \cite{RN848} and MHV \cite{RN697}. MIGD and MHV reflect the overall performance of the algorithms in dynamic environments by averaging the IGD and HV values under multiple environment changes, respectively. They are described in detail in the supplementary material.

\subsection{Results and Analysis on Benchmark Problems}
Table \ref{tab: MIGD of comparison algorithms} shows the mean and standard deviation of the MIGD of the DD-DMOEA algorithm and the comparison algorithms on the 14 test functions. The smaller the MIGD value, the better the algorithm performs. Bolded values in Table \ref{tab: MIGD of comparison algorithms} indicate that the algorithm obtains the optimal MIGD value among all comparison algorithms. To further evaluate the significance of the performance difference between DD-DMOEA and the comparison algorithms, the Wilcoxon rank-sum test is used in this paper \cite{RN403}. In Table \ref{tab: MIGD of comparison algorithms}, the symbols ``+/=/-'' indicate that the MIGD values of DD-DMOEA are significantly better, approximately equal, or significantly worse than the comparison algorithms, respectively.

The data analysis in Table \ref{tab: MIGD of comparison algorithms} reveals that DD-DMOEA significantly outperforms all the compared algorithms under four dynamic environment changes, i.e., DF6, DF10, and DF14. This result is closely related to the method proposed in this paper. Taking DF6 and DF10 as an example, DF6 is a complex problem containing knee points and long-tailed characteristics. The AKP strategy adopted in this paper effectively predicts the knee point changes on the POFs, which enables the DDM to capture more non-dominated solutions on the POFs. The DDM method proposed in this paper effectively alleviates the problem of inequitable solution distribution in DF10 by taking the predicted knee point as the target point and guiding the history obtained POS to approach different target regions gradually. On test functions such as DF1, DF2, DF7, DF9, DF11, and DF13, although DD-DMOEA did not achieve a dominant advantage in all dynamic environments, its performance remained highly competitive and was nearly on par with the best-performing algorithms. For instance, on DF2, DF9, and DF13 under the (5,10) environment, as well as on DF1 and DF7 under the (5,5) environment, DD-DMOEA demonstrated near-optimal performance. 

\begin{table*}[ht!]
\begin{center}
    \caption{Mean and standard deviation of MIGD obtained by the comparison algorithms in four dynamic environments.}
    \label{tab: MIGD of comparison algorithms}
    \resizebox{\textwidth}{!}{
    \begin{tabular}{lccccccccc}
    \toprule
     Prob. & ($n_t,\tau_t$)  & KT-DMOEA  & IGP-DMOEA  & DIP-DMOEA & KTM-DMOEA  & PSGN  & VARE & DM-DMOEA & DD-DMOEA  \\
    \midrule
    \multirow{4}{*}{DF1}   
    & (10, 10) & 0.0402±9.53e-03(+) & 0.0248±3.59e-03(+) & 0.0264±2.27e-03(+) & 0.0242±2.03e-03(+) & 0.0497±5.53e-03(+) & 0.0360±5.01e-03(+) & 0.0304±3.55e-03(+) & \textbf{0.0227±6.70e-04} \\
    & (5, 10) & 0.0549±1.18e-02(+) & \textbf{0.0281±4.23e-03(-)} & 0.0400±3.18e-03(+) & 0.0363±3.15e-03(+) & 0.0496±5.64e-03(+) & 0.0395±5.31e-03(+) & 0.0343±3.79e-03(=) & 0.0348±5.90e-04 \\
    & (10, 5) & 0.1241±2.96e-02(+) & 0.0704±8.84e-03(+) & 0.0665±7.10e-03(+) & 0.0635±7.65e-03(+) & 0.0687±6.94e-03(+) & 0.1285±1.16e-02(+) & 0.1155±6.37e-03(+) & \textbf{0.0544±2.65e-03} \\
    & (5, 5) & 0.1609±3.40e-02(+) & 0.0849±1.19e-02(+) & 0.1086±8.80e-03(+) & \textbf{0.0817±5.78e-03(=)} & 0.0840±9.32e-03(+) & 0.1361±1.45e-02(+) & 0.1246±1.13e-02(+) & 0.0820±3.06e-03 \\
    \midrule
    \multirow{4}{*}{DF2}    
    & (10, 10) & 0.0383±7.69e-03(+) & 0.0275±4.63e-03(+) & 0.0291±3.28e-03(+) & 0.0275±2.69e-03(+) & 0.0455±4.93e-03(+) & 0.0385±3.45e-03(+) & 0.0338±5.20e-03(+) & \textbf{0.0227±8.83e-04} \\
    & (5, 10) & 0.0465±1.12e-02(+) & \textbf{0.0320±5.41e-03(+)} & 0.0413±4.82e-03(+) & 0.0365±3.89e-03(+) & 0.0470±5.54e-03(+) & 0.0402±4.12e-03(+) & 0.0368±6.13e-03(+) & 0.0318±1.32e-03 \\
    & (10, 5) & 0.1135±2.10e-02(+) & 0.0758±9.67e-03(+) & 0.0769±9.86e-03(+) & 0.0760±7.11e-03(+) & \textbf{0.0526±7.28e-03(-)} & 0.1088±8.54e-03(+) & 0.0975±4.71e-03(+) & 0.0623±1.41e-03 \\
    & (5, 5) & 0.1252±2.79e-02(+) & 0.0929±9.48e-03(+) & 0.1122±1.42e-02(+) & 0.0967±8.85e-03(+) & \textbf{0.0559±5.75e-03(-)} & 0.1176±1.02e-02(+) & 0.1063±4.01e-03(+) & 0.0839±1.75e-03 \\
    \midrule
    \multirow{4}{*}{DF3}    
    & (10, 10) & 0.1961±1.30e-02(+) & 0.0887±8.50e-03(+) & 0.0695±9.63e-03(+) & 0.1254±1.24e-02(+) & 0.2453±3.02e-02(+) & 0.1787±1.69e-02(+) & 0.1629±8.52e-03(+) & \textbf{0.0689±5.79e-03} \\
    & (5, 10) & 0.2011±1.62e-02(+) & 0.1178±1.44e-02(+) & \textbf{0.0885±1.48e-02(-)} & 0.1343±1.21e-02(+) & 0.2590±3.03e-02(+) & 0.1889±1.97e-02(+) & 0.1625±9.20e-03(+) & 0.1162±9.33e-03 \\
    & (10, 5) & 0.2885±2.21e-02(+) & 0.1775±1.16e-02(=) & \textbf{0.1354±2.38e-02(-)} & 0.2445±3.06e-02(+) & 0.4191±2.91e-02(+) & 0.2820±3.22e-02(+) & 0.2370±6.82e-03(+) & 0.1853±7.71e-03 \\
    & (5, 5) & 0.2902±2.65e-02(+) & 0.2085±1.52e-02(=) & \textbf{0.1832±2.70e-02(-)} & 0.2682±2.79e-02(+) & 0.4323±4.53e-02(+) & 0.3002±2.57e-02(+) & 0.2698±2.45e-02(+) & 0.2091±4.69e-03 \\
    \midrule
    \multirow{4}{*}{DF4}    
    & (10, 10) & 0.1102±4.76e-03(-) & \textbf{0.1057±4.23e-03(-)} & 0.1527±4.02e-03(+) & 0.1556±4.45e-03(+) & 0.2805±8.29e-02(+) & 0.1290±8.83e-03(-) & 0.1208±7.99e-03(-) & 0.1471±1.55e-03 \\
    & (5, 10) & 0.1107±6.44e-03(-) & \textbf{0.1080±4.46e-03(-)} & 0.2223±5.30e-03(+) & 0.2214±6.28e-03(+) & 0.2186±5.12e-02(+) & 0.1274±1.02e-02(-) & 0.1182±1.13e-02(-) & 0.2136±1.94e-03 \\
    & (10, 5) & 0.2354±2.98e-02(+) & \textbf{0.1400±1.64e-02(-)} & 0.1952±1.96e-02(+) & 0.1975±1.43e-02(+) & 0.4100±9.57e-02(+) & 0.3306±3.42e-02(+) & 0.3014±1.11e-02(+) & 0.1666±8.11e-03 \\
    & (5, 5) & 0.2288±2.79e-02(=) & \textbf{0.1493±2.14e-02(-)} & 0.2629±1.72e-02(+) & 0.2548±1.45e-02(+) & 0.5113±2.10e-01(+) & 0.3333±2.90e-02(+) & 0.3033±2.22e-02(+) & 0.2383±7.22e-03 \\
    \midrule
    \multirow{4}{*}{DF5}    
    & (10, 10) & 0.0353±1.25e-02(+) & \textbf{0.0167±1.97e-03(-)} & 0.0247±9.21e-04(=) & 0.0245±1.09e-03(=) & 0.1476±6.10e-02(+) & 0.0397±5.63e-03(+) & 0.0332±2.46e-03(+) & 0.0250±3.01e-04 \\
    & (5, 10) & 0.0346±7.75e-03(+) & \textbf{0.0183±2.47e-03(-)} & 0.0286±2.66e-03(=) & 0.0277±1.14e-03(=) & 0.1083±3.40e-02(+) & 0.0449±4.29e-03(+) & 0.0404±3.09e-03(+) & 0.0290±4.50e-04 \\
    & (10, 5) & 0.1690±4.91e-02(+) & 0.0447±8.51e-03(=) & \textbf{0.0423±3.39e-03(-)} & 0.0448±5.99e-03(=) & 0.1337±2.86e-02(+) & 0.1722±2.44e-02(+) & 0.1495±5.44e-03(+) & 0.0456±2.51e-03 \\
    & (5, 5) & 0.1724±4.64e-02(+) & 0.0581±8.52e-03(+) & 0.0580±7.66e-03(+) & \textbf{0.0534±8.70e-03(-)} & 0.1873±5.57e-02(+) & 0.1864±3.07e-02(+) & 0.1613±9.25e-03(+) & 0.0573±1.36e-03 \\
    \midrule
    \multirow{4}{*}{DF6}    
    & (10, 10) & 2.1359±6.76e-01(+) & 4.3660±3.96e-01(+) & 2.4035±1.58e+00(+) & 1.4458±3.74e-01(+) & 5.3558±9.10e-01(+) & 0.9789±2.47e-01(+) & \textbf{0.7046±3.97e-01(=)} & 0.7180±1.78e-01 \\
    & (5, 10) & 1.1366±4.88e-01(+) & 2.2615±8.62e-01(+) & 0.9647±8.54e-01(+) & 0.6685±1.93e-01(+) & 6.1481±8.36e-01(+) & 0.7940±1.85e-01(+) & 0.5859±2.24e-01(+) & \textbf{0.4690±1.75e-02} \\
    & (10, 5) & 3.4883±6.25e-01(+) & 5.4163±1.01e+00(+) & 1.4373±4.46e-01(+) & 1.6876±3.81e-01(+) & 10.8291±1.04e+00(+) & 1.3791±3.17e-01(+) & 1.0263±1.90e-01(+) & \textbf{0.8926±9.76e-02} \\
    & (5, 5) & 2.3256±5.89e-01(+) & 2.8663±7.33e-01(+) & 0.9814±2.51e-01(+) & 1.1175±3.75e-01(+) & 11.2535±1.09e+00(+) & 1.0956±2.39e-01(+) & 0.7961±6.31e-01(+) & \textbf{0.6842±8.37e-02} \\
    \midrule
    \multirow{4}{*}{DF7}    
    & (10, 10) & 0.2656±1.97e-02(+) & 0.2870±1.11e-02(+) & 0.1791±2.63e-02(+) & 1.8246±6.38e+00(+) & 0.2087±3.11e-02(+) & 0.3048±3.05e-02(+) & 0.2604±1.16e-02(+) & \textbf{0.1649±3.91e-02} \\
    & (5, 10) & 0.4433±2.58e-02(+) & 0.4028±1.39e-02(+) & 0.3187±6.47e-02(+) & 1.6607±5.41e+00(+) & \textbf{0.2683±3.80e-02(-)} & 0.4322±3.19e-02(+) & 0.3902±4.84e-02(+) & 0.3013±2.47e-02 \\
    & (10, 5) & 0.3304±2.36e-02(+) & 0.3277±1.34e-02(+) & \textbf{0.2065±2.49e-02(-)} & 1.7421±5.98e+00(+) & 0.2345±3.99e-02(-) & 0.3482±3.84e-02(+) & 0.3001±5.97e-02(=) & 0.3084±1.24e-02 \\
    & (5, 5) & 0.4969±1.75e-02(+) & 0.4289±2.96e-02(+) & 0.3424±5.81e-02(+) & 2.5629±8.84e+00(+) & \textbf{0.3143±4.44e-02(=)} & 0.4613±3.03e-02(+) & 0.4269±4.09e-02(+) & 0.3239±1.50e-02 \\
    \midrule
    \multirow{4}{*}{DF8}    
    & (10, 10) & 0.1449±1.02e-02(+) & 0.1073±1.28e-02(-) & 0.1366±9.49e-03(+) & 0.1670±1.19e-02(+) & \textbf{0.0499±1.16e-02(-)} & 0.1858±1.31e-02(+) & 0.1685±8.58e-03(+) & 0.1256±2.48e-03 \\
    & (5, 10) & 0.1479±1.62e-02(+) & 0.1084±1.17e-02(-) & 0.1532±7.70e-03(+) & 0.1804±1.12e-02(+) & \textbf{0.0497±6.78e-03(-)} & 0.1852±1.54e-02(+) & 0.1649±7.54e-03(+) & 0.1353±3.81e-03 \\
    & (10, 5) & 0.2023±1.46e-02(+) & 0.1096±1.15e-02(-) & 0.1541±9.44e-03(+) & 0.2243±1.50e-02(+) & \textbf{0.0888±1.03e-02(-)} & 0.2187±1.15e-02(+) & 0.2031±2.31e-02(+) & 0.1374±4.25e-03 \\
    & (5, 5) & 0.2044±1.58e-02(+) & 0.1138±1.01e-02(-) & 0.1779±1.38e-02(+) & 0.2530±1.20e-02(+) & \textbf{0.0942±8.39e-03(-)} & 0.2194±1.56e-02(+) & 0.1974±1.66e-02(+) & 0.1594±3.27e-03 \\
    \midrule
    \multirow{4}{*}{DF9}    
    & (10, 10) & 0.1495±2.91e-02(+) & 0.1004±1.76e-02(-) & \textbf{0.0991±1.43e-02(-)} & 0.1130±1.75e-02(+) & 0.1144±1.63e-02(+) & 0.2275±2.41e-02(+) & 0.1980±3.33e-02(+) & 0.1101±1.17e-02 \\
    & (5, 10) & 0.1820±4.00e-02(+) & 0.1353±4.17e-02(+) & 0.1937±1.81e-02(+) & \textbf{0.1205±1.10e-02(=)} & 0.1279±1.11e-02(+) & 0.2156±2.48e-02(+) & 0.1795±2.94e-02(+) & 0.1229±1.02e-02 \\
    & (10, 5) & 0.3728±7.97e-02(+) & 0.2533±3.16e-02(+) & 0.2249±2.17e-02(+) & 0.1957±2.18e-02(+) & 0.2268±2.23e-02(+) & 0.3579±3.26e-02(+) & 0.3171±5.61e-02(+) & \textbf{0.1652±8.51e-03} \\
    & (5, 5) & 0.3906±5.74e-02(+) & 0.3647±5.44e-02(+) & 0.2701±3.03e-02(+) & 0.2324±2.75e-02(+) & 0.2745±3.16e-02(+) & 0.3566±2.88e-02(+) & 0.3183±4.99e-02(+) & \textbf{0.2231±1.18e-02} \\
    \midrule
    \multirow{4}{*}{DF10}    
    & (10, 10) & 0.3417±3.92e-02(+) & 0.5829±2.59e-02(+) & 0.2097±2.89e-02(+) & 0.2945±3.17e-02(+) & 0.2159±9.29e-03(+) & 0.3485±1.75e-02(+) & 0.3238±3.34e-02(+) & \textbf{0.1896±1.79e-02} \\
    & (5, 10) & 0.2898±3.63e-02(+) & 0.6018±1.26e-01(+) & 0.2111±2.74e-02(+) & 0.2360±2.07e-02(+) & 0.1971±9.00e-03(+) & 0.2843±1.24e-02(+) & 0.2703±2.13e-02(+) & \textbf{0.1787±8.12e-03} \\
    & (10, 5) & 0.3481±3.30e-02(+) & 0.6140±4.57e-02(+) & 0.2035±2.61e-02(+) & 0.3103±2.69e-02(+) & 0.2455±2.18e-02(+) & 0.3625±1.73e-02(+) & 0.3377±1.29e-02(+) & \textbf{0.1744±9.38e-03} \\
    & (5, 5) & 0.3093±2.89e-02(+) & 0.6660±1.41e-01(+) & 0.2169±1.98e-02(+) & 0.2751±1.85e-02(+) & 0.2253±1.71e-02(+) & 0.3077±2.08e-02(+) & 0.2831±2.62e-02(+) & \textbf{0.1668±6.29e-03} \\
    \midrule
    \multirow{4}{*}{DF11}    
    & (10, 10) & 0.1474±4.28e-03(-) & \textbf{0.1313±4.71e-03(-)} & 0.1747±4.29e-03(=) & 0.1889±3.75e-03(+) & 0.1968±1.58e-02(+) & 0.1440±4.26e-03(-) & 0.1385±2.51e-03(-) & 0.1766±2.08e-03 \\
    & (5, 10) & 0.1462±5.07e-03(-) & \textbf{0.1331±5.83e-03(-)} & 0.2538±4.84e-03(=) & 0.2687±3.45e-03(+) & 0.1981±1.39e-02(-) & 0.1429±4.23e-03(-) & 0.1371±4.29e-03(-) & 0.2541±2.62e-03 \\
    & (10, 5) & 0.1947±6.68e-03(+) & 0.1798±1.14e-02(+) & 0.1706±7.04e-03(+) & 0.1819±7.37e-03(+) & 0.2151±1.29e-02(+) & 0.1883±9.50e-03(+) & 0.1762±8.53e-03(+) & \textbf{0.1598±2.90e-03} \\
    & (5, 5) & 0.1930±8.62e-03(-) & 0.1826±7.75e-03(-) & 0.2495±5.34e-03(+) & 0.2686±6.71e-03(+) & 0.2562±1.58e-02(+) & 0.1813±8.40e-03(-) & \textbf{0.1699±4.35e-03(-)} & 0.2428±2.21e-03 \\
    \midrule
    \multirow{4}{*}{DF12}    
    & (10, 10) & 0.8691±3.82e-02(+) & 0.8112±3.43e-02(+) & 0.7321±2.53e-02(+) & 0.6997±2.64e-02(+) & \textbf{0.2428±1.78e-02(-)} & 0.9152±2.65e-02(+) & 0.8841±1.56e-02(+) & 0.6567±4.82e-03 \\
    & (5, 10) & 0.8137±3.85e-02(+) & 0.8280±3.48e-02(+) & 0.8037±2.86e-02(+) & 0.7044±2.52e-02(+) & \textbf{0.2903±1.76e-02(-)} & 0.9000±2.68e-02(+) & 0.8610±2.07e-02(+) & 0.6719±4.41e-03 \\
    & (10, 5) & 0.9069±2.27e-02(+) & 0.7981±3.90e-02(+) & 0.7955±7.00e-02(+) & 0.8076±4.94e-02(+) & \textbf{0.3107±4.46e-02(-)} & 0.9494±2.57e-02(+) & 0.9173±2.06e-02(+) & 0.6670±9.73e-03 \\
    & (5, 5) & 0.8606±3.00e-02(+) & 0.7986±7.75e-02(+) & 0.8514±5.73e-02(+) & 0.8259±4.41e-02(+) & \textbf{0.3730±6.29e-02(-)} & 0.9415±2.53e-02(+) & 0.9081±1.59e-02(+) & 0.7011±4.42e-03 \\
    \midrule
    \multirow{4}{*}{DF13}    
    & (10, 10) & 0.2618±1.68e-02(+) & 0.2266±9.64e-03(=) & 0.2388±7.81e-03(+) & 0.2436±7.77e-03(+) & 0.3011±1.89e-02(+) & 0.2420±1.43e-02(+) & \textbf{0.2248±4.91e-03(=)} & 0.2322±2.44e-03 \\
    & (5, 10) & 0.2550±1.38e-02(+) & \textbf{0.2293±9.86e-03(-)} & 0.2490±6.69e-03(+) & 0.2557±1.18e-02(+) & 0.3237±2.13e-02(+) & 0.2453±7.75e-03(+) & 0.2377±7.79e-03(=) & 0.2434±2.39e-03 \\
    & (10, 5) & 0.4095±3.50e-02(+) & 0.2653±1.71e-02(+) & 0.2650±3.09e-02(+) & 0.2687±2.11e-02(+) & 0.4072±3.79e-02(+) & 0.3304±1.66e-02(+) & 0.3144±1.00e-02(+) & \textbf{0.2390±5.91e-03} \\
    & (5, 5) & 0.3991±3.64e-02(+) & 0.2811±1.86e-02(+) & 0.2832±2.98e-02(+) & 0.2840±2.22e-02(+) & 0.3797±4.01e-02(+) & 0.3520±2.01e-02(+) & 0.3268±4.72e-03(+) & \textbf{0.2751±5.63e-03} \\
    \midrule
    \multirow{4}{*}{DF14}    
    & (10, 10) & 0.0897±4.47e-03(+) & 0.0858±4.36e-03(+) & 0.0787±2.40e-03(+) & 0.0889±4.44e-03(+) & 1.6839±4.99e-01(+) & 0.0880±4.38e-03(+) & 0.0825±2.16e-03(+) & \textbf{0.0739±1.54e-03} \\
    & (5, 10) & 0.0876±4.37e-03(+) & \textbf{0.0830±4.89e-03(=)} & 0.0898±3.08e-03(+) & 0.0936±4.70e-03(+) & 1.9192±4.89e-01(+) & 0.0882±7.04e-03(+) & 0.0798±2.86e-03(-) & 0.0850±1.87e-03 \\
    & (10, 5) & 0.1771±2.01e-02(+) & 0.1209±5.98e-03(+) & 0.0968±4.83e-03(+) & 0.1175±7.46e-03(+) & 1.4511±3.05e-01(+) & 0.1433±8.65e-03(+) & 0.1325±5.13e-03(+) & \textbf{0.0942±4.10e-03} \\
    & (5, 5) & 0.1819±2.69e-02(+) & 0.1260±6.31e-03(+) & 0.1215±7.32e-03(+) & 0.1285±1.07e-02(+) & 1.5515±4.21e-01(+) & 0.1549±7.72e-03(+) & 0.1431±6.23e-03(+) & \textbf{0.1159±4.16e-03} \\
    \midrule
     \multicolumn{2}{c}{+/=/-} & 50/1/5 & 35/5/16 & 46/4/6 & 50/5/1 & 42/1/13 & 51/0/5 & 45/5/6 & - \\
    \bottomrule
    \end{tabular}%
    }
\end{center}
\end{table*}

Nonetheless, in the DF3, DF4, DF5, DF8, and DF12 test functions, DD-DMOEA did not achieve optimal performance under all four dynamic environments. However, a closer observation reveals that its performance on DF3, DF5, and DF12 remains competitive, second only to the optimal algorithm. On these test functions, DIP-DMOEA and IGP-DMOEA achieved optimal performance in all environments on DF3 and DF4, respectively. The success of DIP-DMOEA is attributed to its ANN-based learning mechanism, which perfectly matches the complex, nonlinear mapping in DF3. The neural network, as a powerful nonlinear function approximator, successfully captures this evolutionary pattern. Secondly, IGP-DMOEA achieved optimal performance on DF4 because it uses its sampling mechanism to generate the desired distribution on the new POF shape directly, and then utilizes the IGP model to find the corresponding decision variables via inverse mapping. This mechanism gives it a unique advantage in handling drastic changes in POF geometry. Meanwhile, the performance advantage shown by PSGN on DF8 and DF12 is attributed to its reinforcement learning-based adaptive search mechanism, which enables it to learn and adapt to the specific characteristics of the problems. On DF8, the core challenge is that the population centroid remains unchanged while the POF morphology changes drastically. PSGN's RL mechanism can learn this characteristic by penalizing ineffective global drift and rewarding effective local morphology reshaping, thus achieving efficient adaptation. In contrast, DD-DMOEA's AKP+DDM is an active guidance strategy designed to migrate the population, which, on DF8, where the centroid should remain unchanged, instead causes unnecessary perturbations. On DF12, facing holes in the POF, PSGN's RL mechanism can quickly adapt to this discontinuous POF by penalizing search behaviors that fall into the holes. However, DD-DMOEA relies on a continuous probability density model to fit the discontinuous true distribution of DF12, leading to model mismatch and performance degradation.
\begin{figure*}[htp]
    \centering
    \includegraphics[width=1\linewidth]{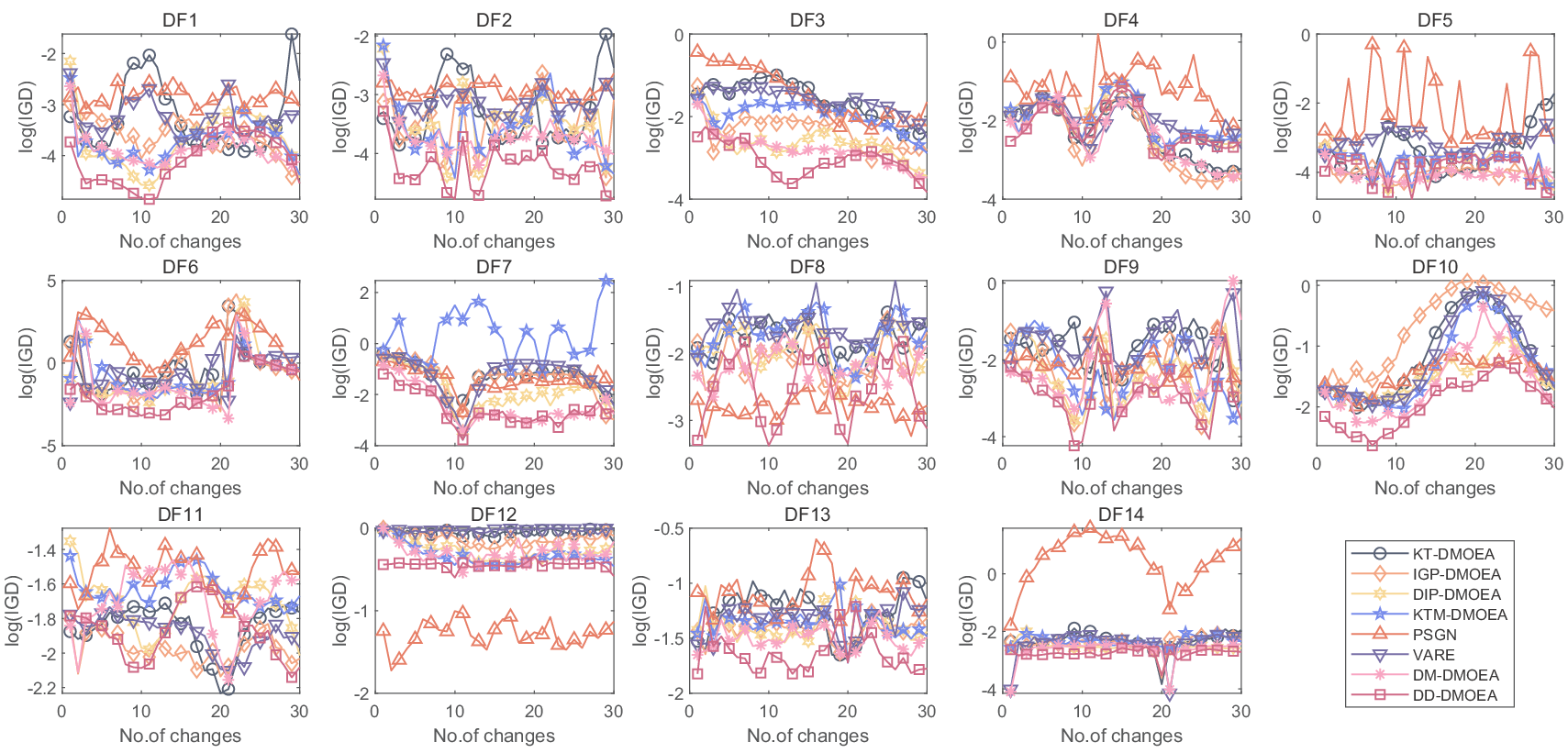}
    \caption{IGD curves for all algorithms on 14 test functions on dynamic environments ($n_t=10$ and $\tau_t=10$).}
    \label{fig: 1-nt10-tauT10-IGD}
\end{figure*}

Due to paper length limitations, the mean and standard deviation of MHV for DD-DMOEA and the comparison algorithms on the 14 test functions are detailed in Table S-II of the supplementary file. A larger MHV value indicates better algorithm performance. As can be seen from Table S-II, the MHV performance of the various algorithms on the 14 test functions is similar to that of MIGD. Although DD-DMOEA did not exhibit optimal MHV performance on all environments for some test functions (such as DF3, DF5, DF7, and DF13), according to the ``+/=/-" statistical results, the comprehensive performance of DD-DMOEA is still superior to all comparison algorithms. Furthermore, to more intuitively display the 30 dynamic change processes of IGD for all algorithms in different dynamic environments, we show the change processes for the four dynamic environments in Fig. \ref{fig: 1-nt10-tauT10-IGD} and Figs. S-1 to S-3 of the supplementary file, respectively, with their dynamic parameters ($n_t$, $\tau_t$) set to (10,10), (5,10), (10,5), and (5,5), respectively. It is worth noting that the vertical axes of these figures use a logarithmic scale to more clearly observe the IGD change trends of the different algorithms. Taking Fig. \ref{fig: 1-nt10-tauT10-IGD} as an example, the IGD curve of DD-DMOEA is at the bottom for most test functions, showing significant advantages, especially on DF2, DF3, DF7, DF10, DF13, and DF14, which indicates that DD-DMOEA can effectively track the POS or POF when dealing with complex and varied environments. However, in the IGD dynamic change process plot for DF12, we can clearly observe that PSGN has a significant advantage over the other algorithms. This is because the penalty mechanism set by PSGN can quickly adapt to the discontinuous POF in the DF12 function, and this is also consistent with the data in Table \ref{tab: MIGD of comparison algorithms}. 

Finally, according to the no free lunch theorem \cite{NFL2018}, the data in Table \ref{tab: MIGD of comparison algorithms} and Table S-II are inherently reasonable, which indicates that no single algorithm exists that is universally applicable to all problems. Although DD-DMOEA failed to show significant performance advantages on some test functions, we believe that the algorithm still possesses great potential. Based on the analysis of the ``+/=/-'' statistical results in Table \ref{tab: MIGD of comparison algorithms} and Table S-II, DD-DMOEA performs the best in terms of comprehensive performance. This advantage is particularly attributed to its novel ``training-free'' diffusion mechanism, which, by explicitly modeling the intermediate transitional states, enables it to more accurately capture the temporal continuity and smooth trends of the POS evolution, thus demonstrating excellent effectiveness in tackling complex DMOPs.

\subsection{Running Time Analysis}
In DMOPs, the algorithm's response speed to dynamic environmental changes is one of the key indicators for evaluating its performance. To evaluate the dynamic response speed of each algorithm, this study compares the dynamic response times of different algorithms under four environmental parameter settings. Due to paper page limitations, this chapter conducts a comparative analysis of the dynamic response times on some test functions, with the results shown in Fig. \ref{fig: rt-1-nt10-tauT10}. Detailed dynamic response time results are presented in Figs. S-8 to S-11 of the supplementary material. Considering the significant differences in dynamic response times among the algorithms, the vertical axis in the figure uses a logarithmic scale to display the comparison results more clearly, while the specific running time values are labeled at the top of each algorithm's bar chart.

Based on the overall comparison results, the dynamic response speeds of the four algorithms, IGP-DMOEA, DIP-DMOEA, KTM-DMOEA, and DM-DMOEA, are relatively slow, which is attributed to their need to execute complex and time-consuming model training or construction processes every time the environment changes. For example, DIP-DMOEA relies on iterative neural network backpropagation training, while KTM-DMOEA employs complex techniques such as iterative boosting or multi-stage manifold transfer learning. Obviously, DM-DMOEA has the longest dynamic response time. This is because DM-DMOEA employs a traditional diffusion model for prediction, and even though a VAE is adopted to reduce dimensionality, the training of deep learning models inherently still incurs high computational costs.

In contrast, PSGN, VARE, and the proposed DD-DMOEA all adopt computationally more efficient analytical or training-free methods. Specifically, although PSGN uses a neural network, its training relies on the incremental ELM, which is a fast method that determines weights in one step through analytical computation. VARE similarly relies on a fast analytical statistical model and is further accelerated by PCA dimensionality reduction. Under the four environmental parameter settings, although the dynamic response speed of the DD-DMOEA proposed in this paper is not always the fastest, it is nearly on par with that of the fastest algorithms. This demonstrates that the training-free diffusion mechanism adopted in this paper can respond more quickly to dynamic environmental changes compared to most current prediction-based methods.
\begin{figure}[t]
    \centering
    \includegraphics[width=\linewidth]{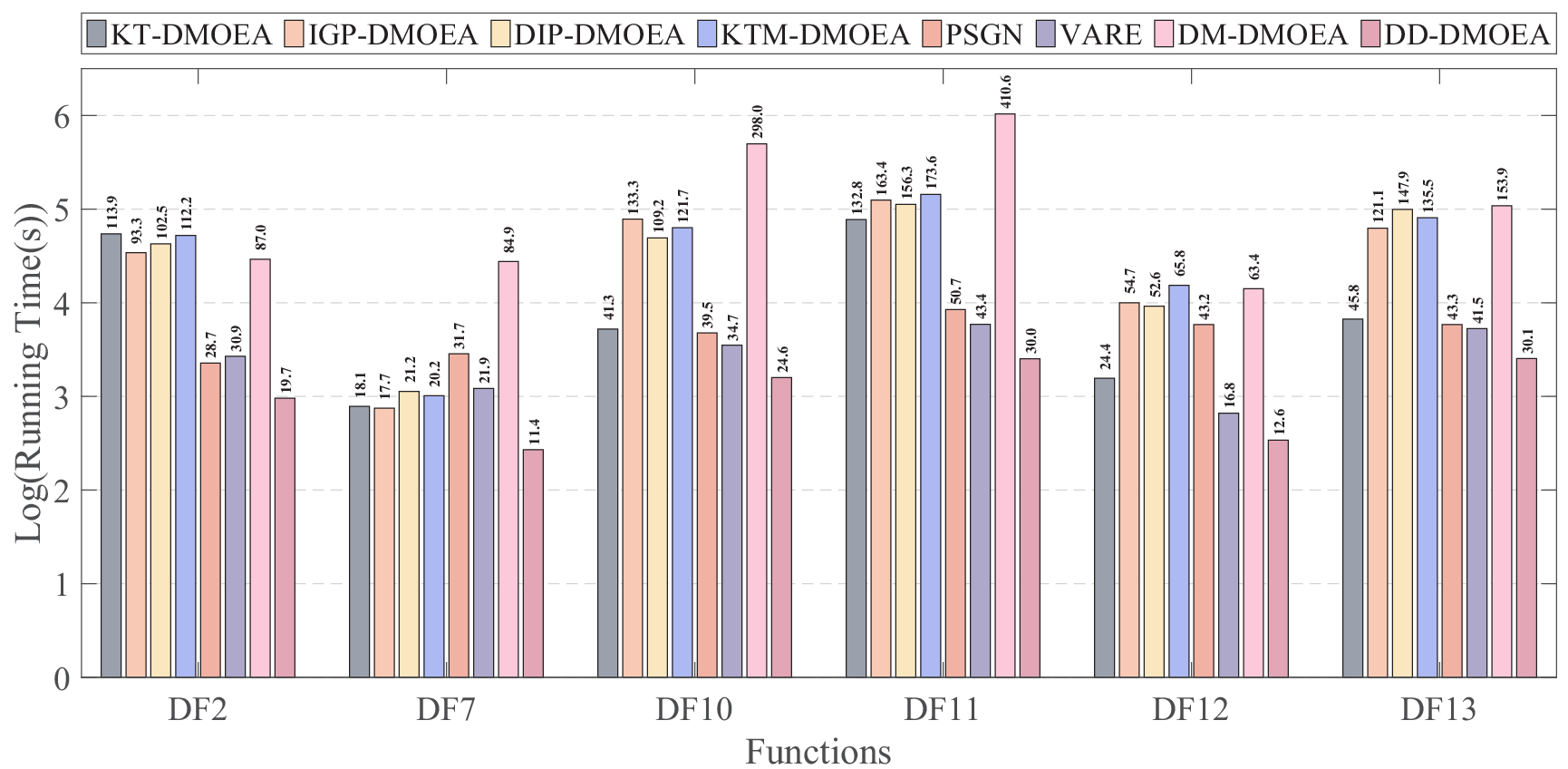}
    \caption{Average running time of all algorithms on some test functions on dynamic environments ($n_t=10$ and $\tau_t=10$).}
    \label{fig: rt-1-nt10-tauT10}
\end{figure}

\subsection{Ablation Study}
To evaluate the effectiveness of the AKP strategy and the DDM strategy, three comparison algorithms, DD-DMOEA/$\text{v}_1$, DD-DMOEA/$\text{v}_2$, and DD-DMOEA/$\text{v}_3$, are designed to compare their performance with the DD-DMOEA algorithm proposed in this paper, respectively. Specifically, DD-DMOEA/$\text{v}_1$ replaces the AKP strategy with the classical linear prediction method of knee points, aiming at verifying the superiority of the AKP strategy in dynamic environments, while DD-DMOEA/$\text{v}_2$ and DD-DMOEA/$\text{v}_3$ replace the DDM strategy with the historical POS and the random individuals, respectively, to evaluate the performance of the DDM strategy in generating the performance advantages in generating high-quality solutions. Their MIGD and MHV comparison results are detailed in Tables S-III and S-IV of the supplementary material. The results of the ``+/=/-'' statistics show that DD-DMOEA significantly outperforms the three compared algorithms in comprehensive performance. Among them, the experimental results of DD-DMOEA/$\text{v}_1$ confirm the limitations of the classical linear prediction knee point method in dealing with various dynamic environmental changes. In addition, the experimental results of DD-DMOEA/$\text{v}_2$ and DD-DMOEA/$\text{v}_3$ show that the use of both historical POS and random individuals negatively affects the algorithm's performance. Notably, DD-DMOEA/$\text{v}_2$ slightly underperforms DD-DMOEA/$\text{v}_3$, this phenomenon indicates that relying only on historical POS, although it can retain some high-quality solutions, is difficult to maintain population diversity and is only effective for problems with continuous environmental similarities.

\subsection{Parameter Sensitivity Analysis}
\label{Parameter Sensitivity Analysis}
\begin{figure}[t]
    \centering
    \includegraphics[width=\linewidth]{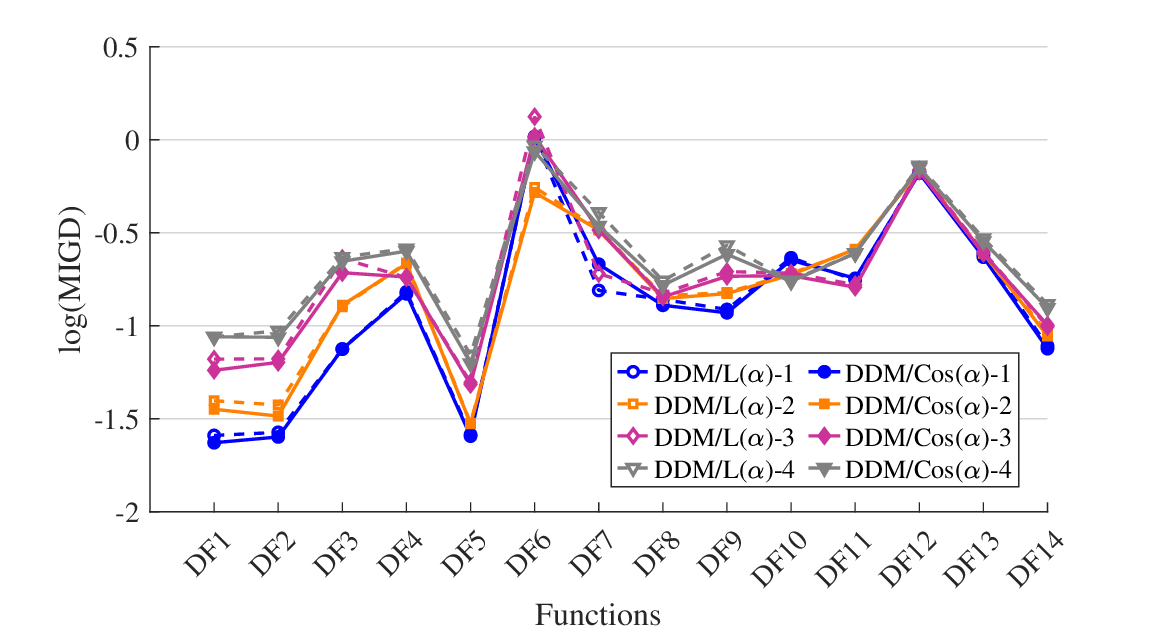}
    \caption{MIGD values generated by different noise schedules $\alpha$ for all test functions in four dynamic environments. For example, DDM/L($\alpha$)-1 and DDM/C($\alpha$)-1 denote the linear and cosine schedules on the dynamic environment ($n_t=10$ and $\tau_t=10$).}
    \label{fig: diff-alpha}
\end{figure}
\begin{figure}[t]
    \centering
    \includegraphics[width=\linewidth]{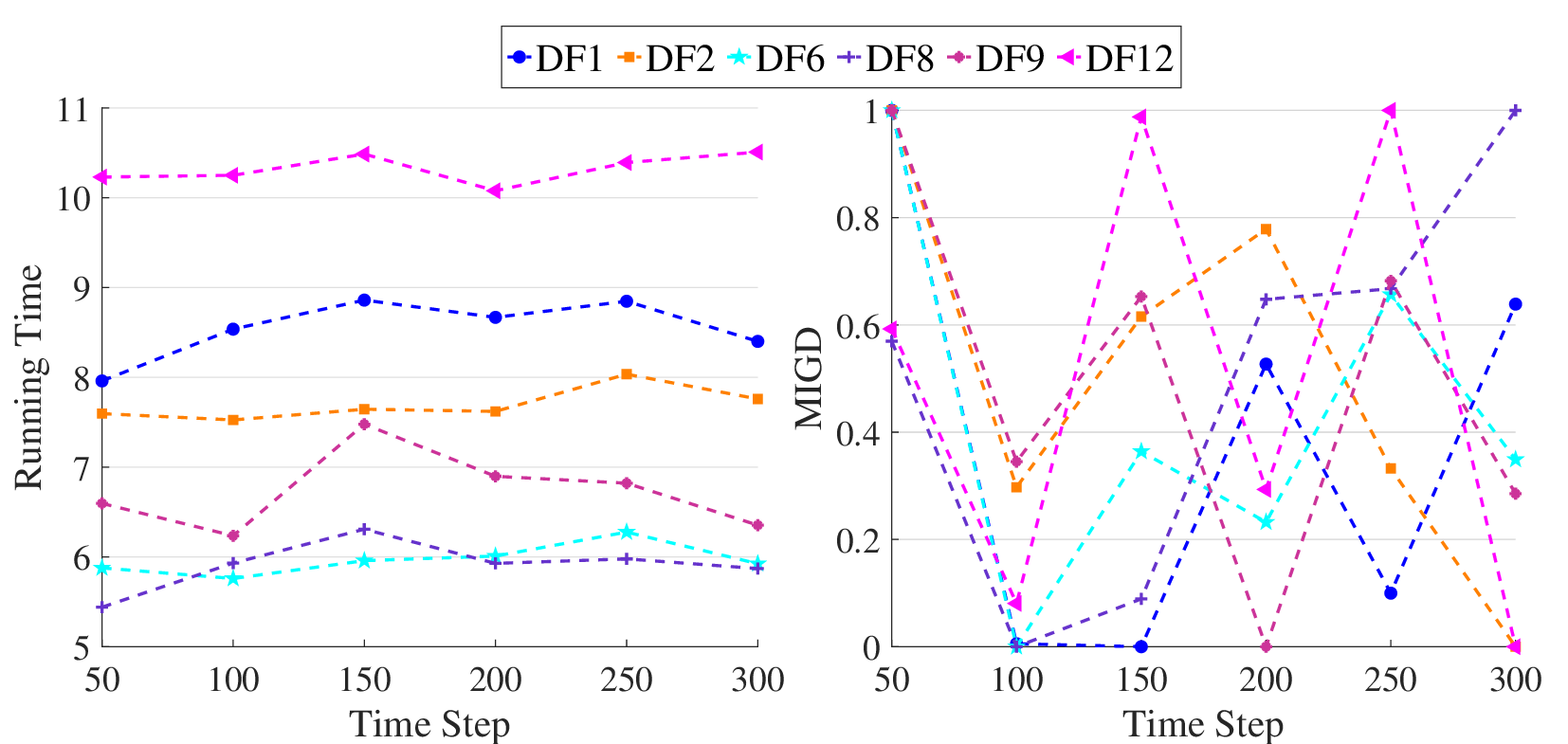}
    \caption{Average running time and MIGD value for six test functions at different time steps on dynamic environments ($n_t=10$ and $\tau_t=5$).}
    \label{fig: Parsen-3-nt10-tauT5}
\end{figure}

In the DDM method, the noise scheduling parameter $\alpha$, the total time step M, and the standard deviation $\psi$ of the Gaussian probability distribution serve as the key factors affecting generation quality, and this paper verifies their influence mechanisms by parameter sensitivity experiments, the results of which are recorded in Tables S-V and S-VII of the supplementary material. For noise scheduling $\alpha$, the comparison experiment is especially set up to compare and analyze the linear scheduling scheme (L$(\alpha)$=1-$k/K$) with the cosine scheduling scheme proposed in this paper. The MIGD comparison curves of the 14 test functions in the four dynamic environments shown in Fig. \ref{fig: diff-alpha} indicate that the curve of DDM/Cos($\alpha$) is always located below that of DDM/L($\alpha$), which verifies that the cosine scheduling has a better environmental adaptation capacity in DMOPs. Regarding the evaluation of the total time step, this study constructs six sets of parameter configurations containing M=[50,100,150,200,250,300], where the comparison algorithms corresponding to $K$=100 used in this paper are labeled DDM/$K_1$ ($K_1$=50) to $K_5$ ($K_5$=300), respectively. The multiple time step comparison results in Fig. \ref{fig: Parsen-3-nt10-tauT5} show that the increase in time steps did not have a significant effect on the algorithm running time. In contrast, when M = 100, MIGD has a small value in most functions, and the MIGD value does not decrease significantly as the time step increases. This proves that setting the time step to 100 ensures both the quality of the solution and computational efficiency. Owing to page limitations in the paper, the results and analysis of the sensitivity experiments for $\psi$ are presented in the supplementary materials.

\begin{table*}[t]
  \centering
  \caption{Mean and standard deviation of HV obtained by the comparison algorithm for five environmental changes in the ROA problem.}
    \label{tab: ROA optimization}%
    \begin{tabular}{cccccc}
    \toprule
     Algs. &  $\text{ENV}_1$ &  $\text{ENV}_2$ &  $\text{ENV}_3$ &  $\text{ENV}_4$ &  $\text{ENV}_5$ \\
     \midrule
    KT-DMOEA & 2.29e-01±(2.11e-02) & 2.14e-01±(9.86e-02) & 1.83e-01±(9.29e-02) & 1.79e-01±(6.59e-02) & 2.14e-01±(5.41e-02) \\
    IGP-DMOEA & 2.02e-01±(9.95e-02) & 2.20e-01±(5.50e-02) & 1.81e-01±(7.00e-02) & 2.03e-01±(4.13e-02) & 1.93e-01±(5.41e-02) \\
    DIP-DMOEA & 2.05e-01±(4.32e-02) & 2.02e-01±(3.76e-02) & 1.98e-01±(2.47e-02) & 1.85e-01±(2.07e-02) & 2.01e-01±(2.54e-02) \\
    KTM-DMOEA & 1.99e-01±(4.22e-02) & 2.00e-01±(3.25e-02) & 1.88e-01±(3.31e-02) & 2.02e-01±(3.46e-02) & 2.03e-01±(2.76e-02) \\
    PSGN & 2.57e-01±(1.07e-01) & 2.39e-01±(1.05e-01) & 1.89e-01±(1.05e-01) & \textbf{2.31e-01±(6.41e-02)} & 2.39e-01±(6.58e-02) \\
    VARE & 1.88e-01±(5.47e-02) & 2.06e-01±(7.38e-02) & 1.86e-01±(7.19e-02) & 2.16e-01±(5.80e-02) & 1.88e-01±(3.27e-02) \\
    DM-DMOEA & 1.78e-01±(4.37e-02) & 1.91e-01±(3.22e-02) & 2.15e-01±(3.68e-02) & 1.87e-01±(2.22e-02) & 2.06e-01±(4.50e-02) \\
    DD-DMOEA & \textbf{3.25e-01±(4.39e-02)} & \textbf{3.80e-01±(6.62e-02)} & \textbf{3.26e-01±(4.22e-02)} & 2.24e-01±(9.42e-03) & \textbf{3.03e-01±(3.68e-02)} \\
    \bottomrule
    \end{tabular}%
\end{table*}%

\subsection{Real-world Application}
In this paper, a raw ore allocation (ROA) problem is modeled as a DMOP to evaluate the dynamic optimization capability of DD-DMOEA in real-world applications. The ROA problem is a typical class of optimization problems in the field of mining and resource processing, aiming to optimize the efficiency, cost, and quality of the whole production chain through the scientific allocation of resources \cite{roa2019}. However, it faces the challenge of dynamic changes of input parameters and constraints with the environment and time in real operations, so it is modeled as a DMOP to verify the dynamic responsiveness of the algorithm. Specifically, the optimization objective is mainly to minimize the production cost while increasing the concentrate yield, grade, metal recovery, and reducing the beneficiation ratio. In this experiment, the population size is set to 200, the dimensionality of the decision variables is 6 and the solution space is normalized to the interval [0,1], and each algorithm is iterated 50 times and run independently 20 times when the environment changes in an environment containing 5 dynamic environments ($\text{ENV}_1$-$\text{ENV}_5$). The mean and standard deviation of the HV obtained by DD-DMOEA and the comparison algorithms on the ROA problem are shown in Table \ref{tab: ROA optimization}. The experimental results show that DD-DMOEA achieves optimal performance in four environmental conditions ($\text{ENV}_1$, $\text{ENV}_2$, $\text{ENV}_3$, and $\text{ENV}_5$), while PSGN has a better dynamic response capability on $\text{ENV}_4$. Although DD-DMOEA does not always obtain the optimal solution in all five environments, the overall performance is better than that of all the compared algorithms, thus demonstrating the good dynamic optimization ability of DD-DMOEA for the ROA problem.

\section{Conclusion} 
\label{Conclusion}

This paper has investigated dynamic multiobjective optimization from the perspective of population transition modeling and proposed a training-free diffusion-driven evolutionary algorithm, termed DD-DMOEA. By interpreting the Pareto optimal solution set in consecutive environments as states of a diffusion process, the proposed approach formulates the dynamic response as an analytical multi-step denoising procedure rather than a one-step population mapping. This design allows the algorithm to capture the gradual evolutionary trend of the POS while avoiding the training cost and response delay associated with learning-based predictors.

A knee-point-based auxiliary strategy is introduced to specify the target region in the new environment, and an explicit probability-density formulation is derived to guide the denoising process without neural networks. To improve robustness under complex and irregular environmental changes, an uncertainty-aware guidance mechanism adaptively adjusts the diffusion strength according to historical prediction deviations, thereby reducing the risk of misleading population relocation.

Extensive experimental results on the CEC2018 dynamic multiobjective benchmarks demonstrate that DD-DMOEA provides an effective balance between convergence and diversity while achieving faster response to environmental changes compared with several representative state-of-the-art DMOEAs. These results indicate that analytically driven diffusion mechanisms offer a viable and efficient alternative to learning-based prediction models for dynamic multiobjective optimization.

Future work will focus on extending the proposed diffusion-driven framework to more complex dynamic scenarios, including higher-dimensional objective spaces and problems with heterogeneous or partially observable environmental changes.
Experimental results demonstrate that DD-DMOEA outperforms existing methods in terms of both solution quality and dynamic response speed, making it a highly promising solution for real-time dynamic multiobjective optimization tasks. Future work will explore further improvements in model scalability and its application to more complex, real-world dynamic optimization scenarios.


\bibliography{mybibtex}

\end{document}


\IEEEiedlistdecl
%

%
%
%
%

\title{Supplementary Material of ``Training-Free Diffusion-Driven Modeling of Pareto Set Evolution for Dynamic Multiobjective Optimization''}

 \IEEEtitleabstractindextext{%
\begin{abstract}
This is the supplementary material of ``Training-Free Diffusion-Driven Modeling of Pareto Set Evolution for Dynamic Multiobjective Optimization''. It contains detailed information about the CEC2018 benchmark suites and the performance metrics, MIGD values, and MHV values for DD-DMOEA with comparison algorithms, where the comparison algorithms contain all the algorithms for comparison experiments, ablation experiments, and parameter sensitivity experiments. In addition, this material shows the dynamic change curves of MIGD for DD-DMOEA with comparison algorithms on 14 test functions and the true POFs obtained by all algorithms on some of the test functions.

\end{abstract}
}
\bibliographystyle{IEEEtran}
\maketitle

\IEEEdisplaynontitleabstractindextext

\IEEEpeerreviewmaketitle

\section{INTRODUCTION}
\begin{enumerate}
\item TABLE \ref{tab: test functions}: Dynamic characteristics of 14 test functions in the CEC2018 dynamic multi-objective optimization competition.

\item  TABLE \ref{tab: MHV of comparison algorithms}: Mean and standard deviation of MHV obtained by the comparison algorithms in four dynamic environments.

\item  Fig. \ref{fig: 2-nt5-tauT10-IGD}$\sim$\ref{fig: 4-nt5-tauT5-IGD}: IGD curves for all algorithms on 14 test functions on dynamic environments ($n_t=5$ and $\tau_t=10$), ($n_t=10$ and $\tau_t=5$), and ($n_t=5$ and $\tau_t=5$).


\item  Fig. \ref{fig: rt-1-nt10-tauT10}$\sim$\ref{fig: rt-4-nt5-tauT5}: Average running time of all algorithms on 14 test functions on dynamic environments ($n_t=10$ and $\tau_t=10$), ($n_t=5$ and $\tau_t=10$), ($n_t=10$ and $\tau_t=5$), and ($n_t=5$ and $\tau_t=5$).

\item  TABLE \ref{Ablation results-MIGD}: Mean and standard deviation of MIGD values of all test functions for different strategies on four dynamic environments.

\item  TABLE \ref{Ablation results-MHV}: Mean and standard deviation of MHV values of all test functions for different strategies on four dynamic environments.

\item  TABLE \ref{ParSen results-MIGD}: Mean and standard deviations of MIGD values obtained from all test functions in four dynamic environments by different time steps $M$ and noise scheduling $\alpha$.

\item  TABLE \ref{ParSen results-MHV}: Mean and standard deviations of MHV values obtained from all test functions in four dynamic environments by different time steps $M$ and noise scheduling $\alpha$.

\item TABLE \ref{ParSen results-Std MIGD}: Mean and standard deviations of MIGD values obtained from all test functions in four dynamic environments by the standard deviation $\psi$ of different Gaussian distributions. 

\item TABLE \ref{ParSen results-Std MHV}: Mean and standard deviations of MHV values obtained from all test functions in four dynamic environments by the standard deviation $\psi$ of different Gaussian distributions.
\end{enumerate}

\section{EXPERIMENTS}
\subsection{CEC2018 benchmark suites}
In the CEC2018 dynamic multiobjective optimization benchmark test suite, the researchers constructed an evaluation system containing 14 benchmark functions. The benchmark set effectively simulates the typical scenario characteristics in engineering optimization through the dimensions of time-dependent Pareto front (PF) or solution set (PS) geometric properties, irregular PF shapes, and nonlinear correlation variables. Specifically, the benchmark set consists of nine bi-objective optimization problems and five tri-objective optimization problems, and its core challenges are reflected in three aspects: time-dependent properties of PF/PS geometries, complex topological structures, and non-uniform feature distributions. Table \ref{tab: test functions} systematically summarizes the core dynamic properties of each test problem and their corresponding mathematical representations.

\begin{table*}[htp]
    \centering
    \caption{Dynamic characteristics of 14 test functions in the CEC2018 dynamic multi-objective optimization competition.}
    \label{tab: test functions}
    \resizebox{\textwidth}{!}{
    \begin{tabular}{ccll}
        \toprule
        Problem & objectives &  Dynamics & Remarks \\
        \midrule
        DF1 & 2 & mixed convexity-concavity,location of optima      & dynamic PF and PS \\
        \midrule
        DF2 & 2 & switch of position-related variable, location of optima & static convex PF, dynamic PS, severe diversity loss \\
        \midrule
        DF3 & 2 & mixed convexity-concavity,variable-linkage, location of optima & dynamic PF and PS \\
        \midrule
        DF4 & 2 & variable-linkage, PF range, bounds of PS & dynamic PF and PS \\
        \midrule
        DF5 & 2 & number of knee regions, local of optima  & dynamic PF and PS  \\
        \midrule
        DF6 & 2 & mixed convexity-concavity, multimodality, location of optima  & dynamic PF and PS \\
        \midrule
        DF7 & 2 & PF range, location of optima   & convex PF, static PS centroid, dynamic PF and PS \\
        \midrule
        DF8 & 2 & mixed convexity-concavity, distribution of solutions, location of optima & static PS  centroid, dynamic PF and PS, variable-linkage \\
        \midrule
        DF9 & 2 & number of disconnected PF segments, location of optima & dynamic PS and PF, variable-linkage \\
        \midrule
        DF10 & 3 & mixed convexity-concavity, location of optima & dynamic PS and PF, variable-linkage \\
        \midrule
        DF11 & 3 & size of PF region, PF range, location of optima & dynamic PS and PF, concave PF, variable-linkage \\
        \midrule
        DF12 & 3 & number of PF holes, location of optima & dynamic PS, static concave PF, variable-linkage \\
        \midrule
        \multirow{2}{*}{DF13} & \multirow{2}{*}{3} & \multirow{2}{*}{number of disconnected PF segments, location of optima} & dynamic PS and PF, the PF can be a continuous convex\\ & &  & or concave segment, or several disconnected segments  \\
        \midrule
        DF14 & 3 & degenerate PF, number of knee regions, location of optima & dynamic PS and PF, variable-linkage \\
        \bottomrule
    \end{tabular}
    }
\end{table*}

\subsection{Comparative Algorithms}
The proposed DD-DMOEA was experimentally compared against six state-of-the-art algorithms, including KT-DMOEA, IGP-DMOEA, KTM-DMOEA, DIP-DMOEA, PSGN, VARE, and DM-DMOEA. Among them:
\begin{enumerate}
\item KT-DMOEA first utilizes a trend prediction model to estimate high-quality knee points, and then employs an imbalanced transfer learning technique to generate an initial population using these few knee points.

\item IGP-DMOEA utilizes an inverse Gaussian process to construct a predictor that maps historical optimal solutions from the objective space back to the decision space, thereby generating an effective initial population that better meets the decision-maker's preferences.

\item KTM-DMOEA combines two strategies: ``Knowledge Transfer Prediction" as a discriminative predictor and ``Knowledge Maintenance Sampling" as a generative predictor, to jointly produce a high-quality initial population and alleviate negative transfer.

\item DIP-DMOEA designs a neural network to learn the environment's change patterns and proposes a Directional Improvement Prediction strategy, which combines an improvement term learned by the neural network and a linear difference term to jointly guide the population's evolutionary direction.

\item PSGN proposes a particle search guidance network, which uses reinforcement learning to learn how to guide the search behavior of individuals, and employs an incremental extreme learning machine to reduce computational costs.

\item VARE proposes a method combining vector autoregression and environment-aware hypermutation (EAH), where the VAR model captures the inter-dependencies between decision variables for population prediction, while EAH serves as an adaptive complementary strategy to adjust diversity by sensing changes in both decision and objective spaces when prediction is unsuitable.

\item DM-DMOEA proposes a diffusion model-based prediction strategy that incorporates a variational autoencoder and a ``PS estimation" method, aiming to learn the distribution relationships of solutions by integrating historical and new environmental information, thereby generating high-quality initial populations in complex, irregular, and severe environmental changes.
\end{enumerate}

\subsection{Performance Metric}
\label{supp-Performance Metric}
Two performance metrics are used in this experiment to comprehensively evaluate the convergence and diversity of the algorithms in DMOPs, i.e., MIGD \cite{RN848} and MHV \cite{RN697}. MIGD and MHV reflect the overall performance of the algorithms in dynamic environments by averaging the values of IGD and HV, respectively, over multiple environment changes. Their calculation formulas are as follows:
\begin{equation}
\begin{split}
    \text{MIGD} &= \frac{1}{T} \sum_{t\in T} \text{IGD}(P_t^*, P_t) \\
    \text{MHV} &= \frac{1}{T} \sum_{t\in T} \text{HV}(P_t, r_t)
\end{split},    
\end{equation}
where $T$ is the total number of environmental changes. $P^*$ is the true POF at moment $t$, and $P_t$ is the solution set obtained by SMOEA at moment $t$. $r_t$ is the reference point at moment $t$. $\text{MIGD}(P^*, P_t)$ \cite{IGD} and $\text{HV}(P^*,r_t)$ \cite{HV} then denote the IGD value and HV value calculated at moment $t$, respectively.
\begin{equation}
\begin{split}
    \text{IGD}(P^*, P) &= \frac{1}{|P^*|} \sum_{v \in P^*} \min_{u \in P} d(u, v) \\
    \text{HV}(P, r) &=  \mathcal{L}_m \left( \bigcup_{u \in P} \text{Box}(u, r) \right)
\end{split},    
\end{equation}
where $P$ and $P^*$ are the solution set, and the true POF obtained by the algorithm, respectively, and $\left | P^* \right |$ is the number of solutions in $P^*$. $d(u,v)$ is the Euclidean distance between the solution $u$ and the reference point $v$. $\mathcal{L}_m$ is the Lebesgue measure of the $m$-dimensional space and $m$ is the number of objective functions. $\text{Box}(u, r)$ denotes the hypervolume defined by the solution $u$ and the reference point $r$.

\subsection{Experiments Results and Analysis}
\begin{figure*}
    \centering
    \includegraphics[width=\linewidth]{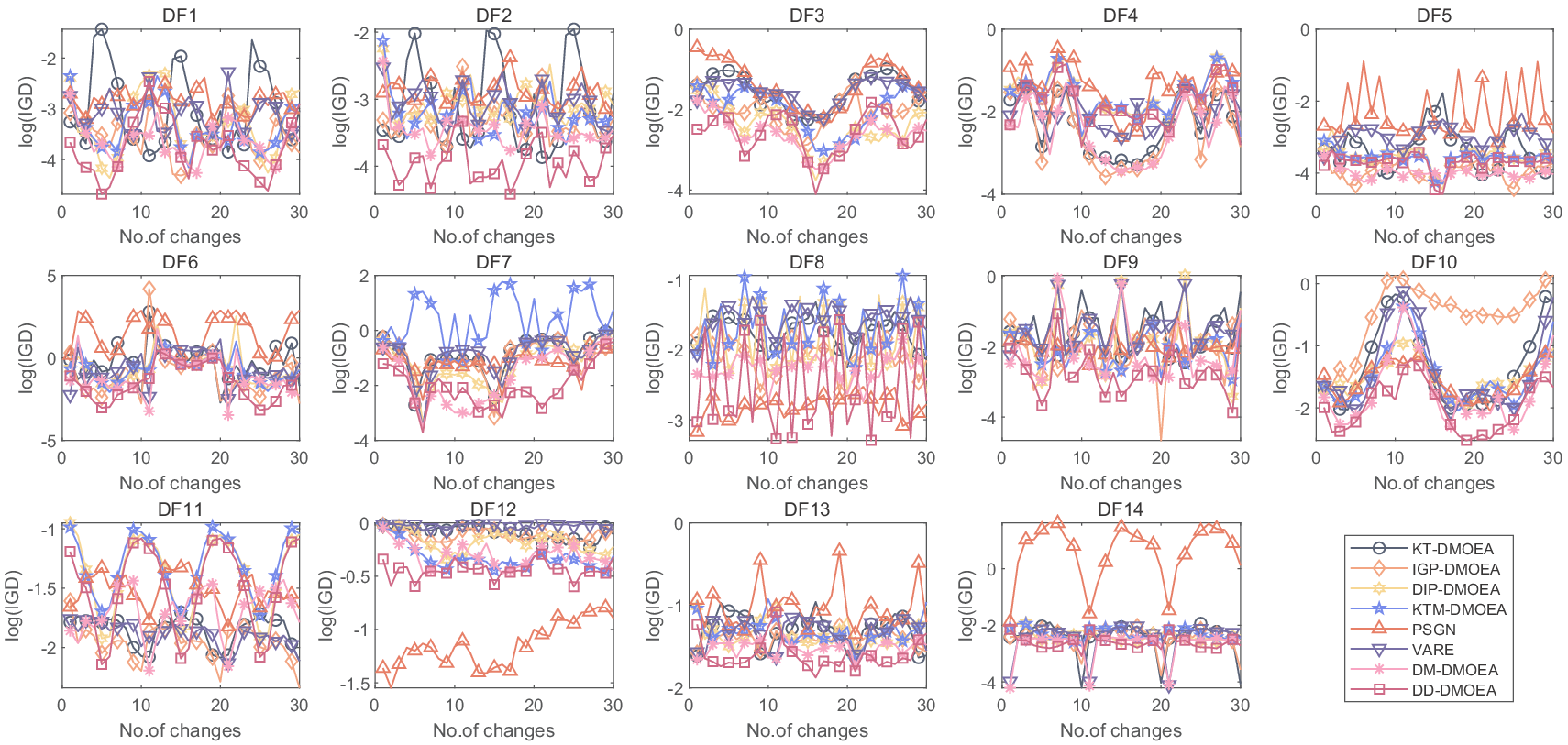}
    \caption{IGD curves for all algorithms on 14 test functions on dynamic environments ($n_t=5$ and $\tau_t=10$).}
    \label{fig: 2-nt5-tauT10-IGD}
\end{figure*}

\begin{figure*}
    \centering
    \includegraphics[width=\linewidth]{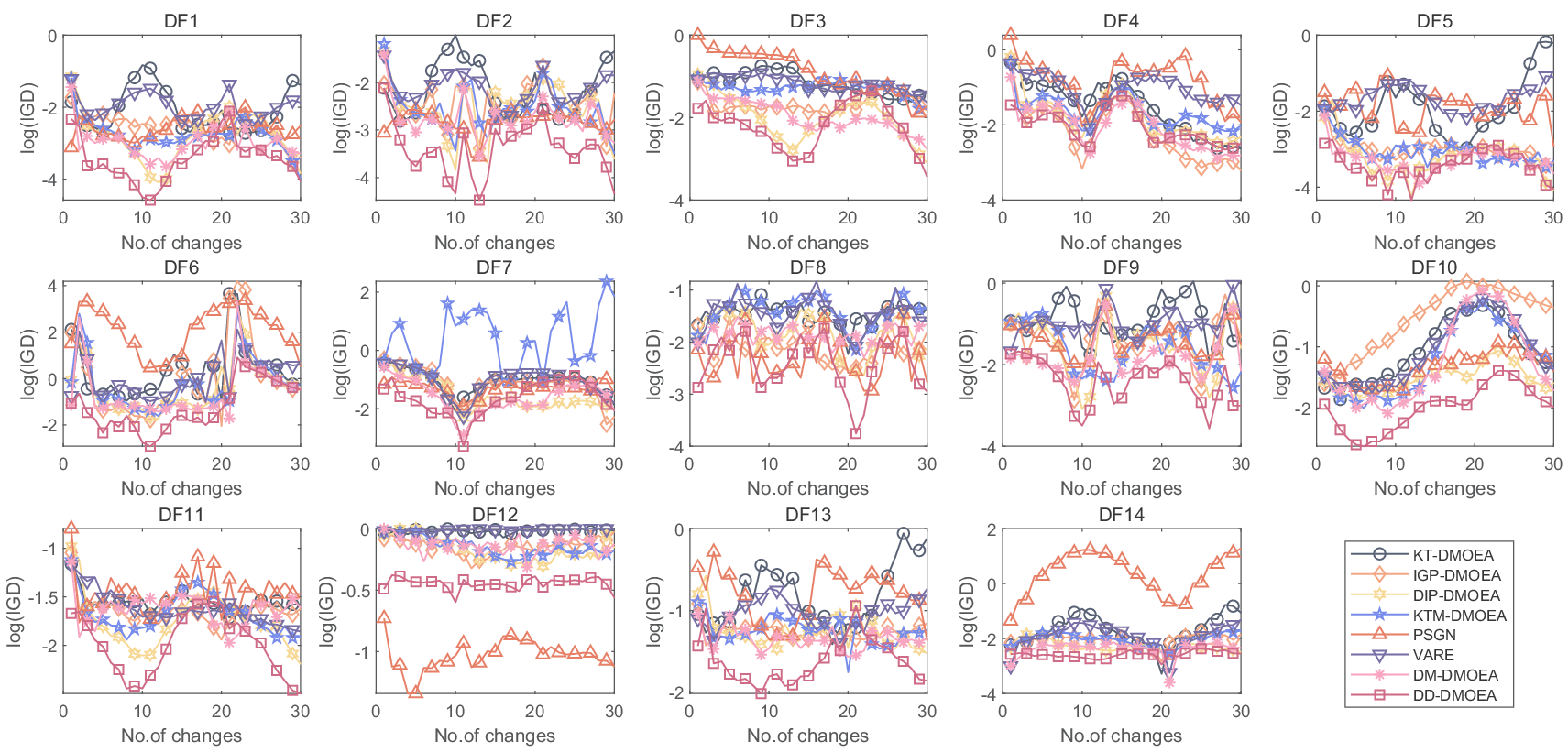}
    \caption{IGD curves for all algorithms on 14 test functions on dynamic environments ($n_t=10$ and $\tau_t=5$).}
    \label{fig: 3-nt10-tauT5-IGD}
\end{figure*}

\begin{figure*}
    \centering
    \includegraphics[width=\linewidth]{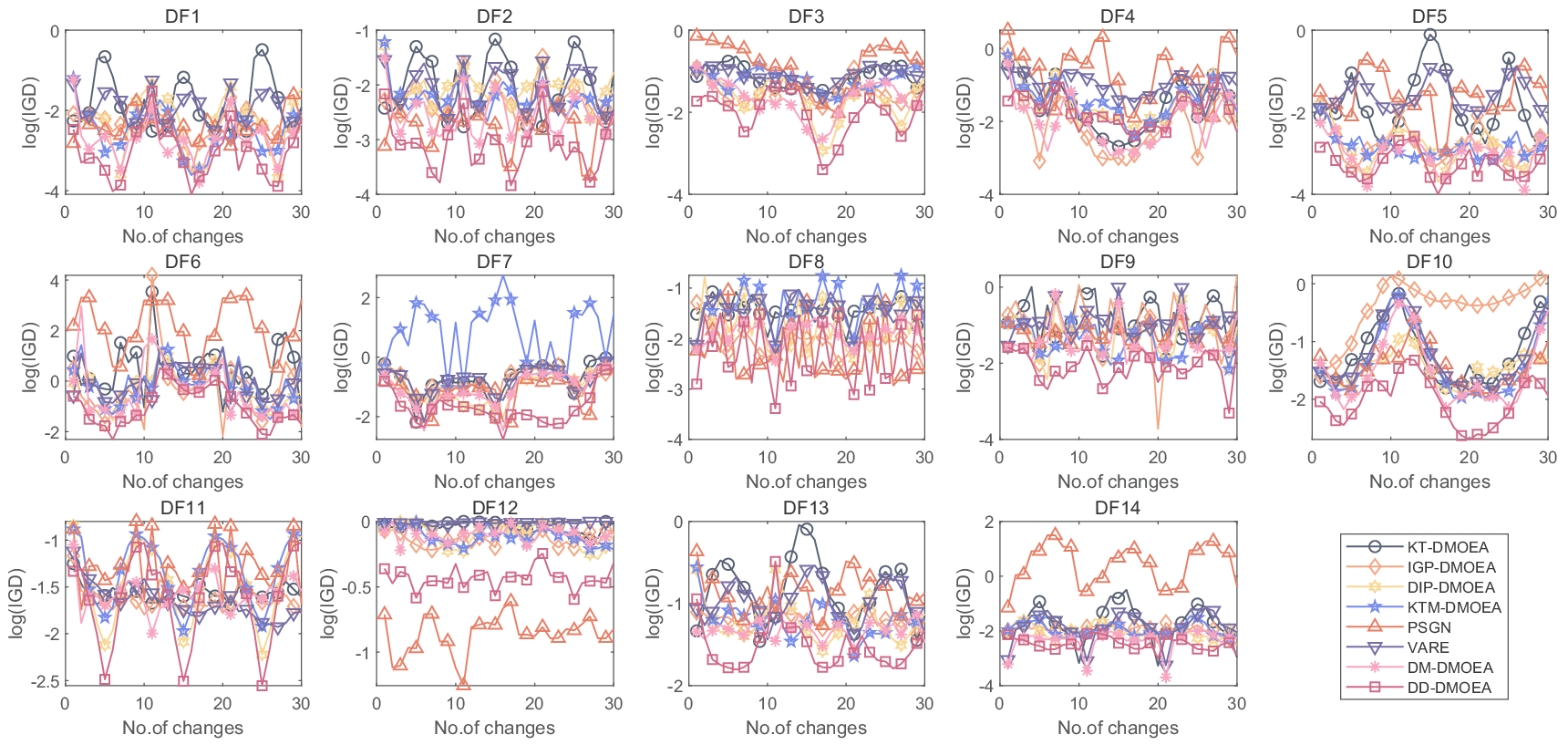}
    \caption{IGD curves for all algorithms on 14 test functions on dynamic environments ($n_t=5$ and $\tau_t=5$).}
    \label{fig: 4-nt5-tauT5-IGD}
\end{figure*}

According to the experimental data in Table \ref{tab: MHV of comparison algorithms}, it can be found that the DD-DMOEA algorithm demonstrates a significant superiority in the test functions with multiple knee points of Pareto front shape (e.g., DF2, DF6, DF8, DF9, DF10, and DF11, etc.). This superiority is mainly attributed to the collaborative effect of its DDM and AKP, that is, the effective construction of high-quality initial populations adapted to dynamic environments through the gradual movement of the POS, a conclusion that is consistent with the analytical results of the MIGD metrics. It is notable that in the test problems DF1(5,10), DF1(5,5), DF9(10,10), and DF12(10,10), although DD-DMOEA did not achieve the optimal performance, it still maintains the front rank level, demonstrating its potential for dynamic adaptation to the environment. Meanwhile, the advantage of DIP-DMOEA on DF3 stems from its ANN-based learning mechanism. The core challenge of DF3 is the complex nonlinear variable linkage, and DIP-DMOEA's ANN, as a powerful nonlinear function approximator, can effectively learn and fit this evolutionary pattern. On the other hand, KTM-DMOEA's optimal performance on DF5 and DF13 is attributed to its knowledge maintenance sampling strategy. KMS generates new solutions by sampling through constructing and perturbing a multivariate Gaussian distribution, and this generative method based on probability distribution makes it more robust when dealing with drastic changes in POF geometry. In addition, Figs \ref{fig: 2-nt5-tauT10-IGD} to \ref{fig: 4-nt5-tauT5-IGD} demonstrate the evolution of the algorithm under different dynamic parameters, in which the MIGD curve of DD-DMOEA maintains a significant advantage in most of the subfigures, reflecting the superiority of the algorithm in DMOPs.

\begin{figure*}
    \centering
    \includegraphics[width=\linewidth]{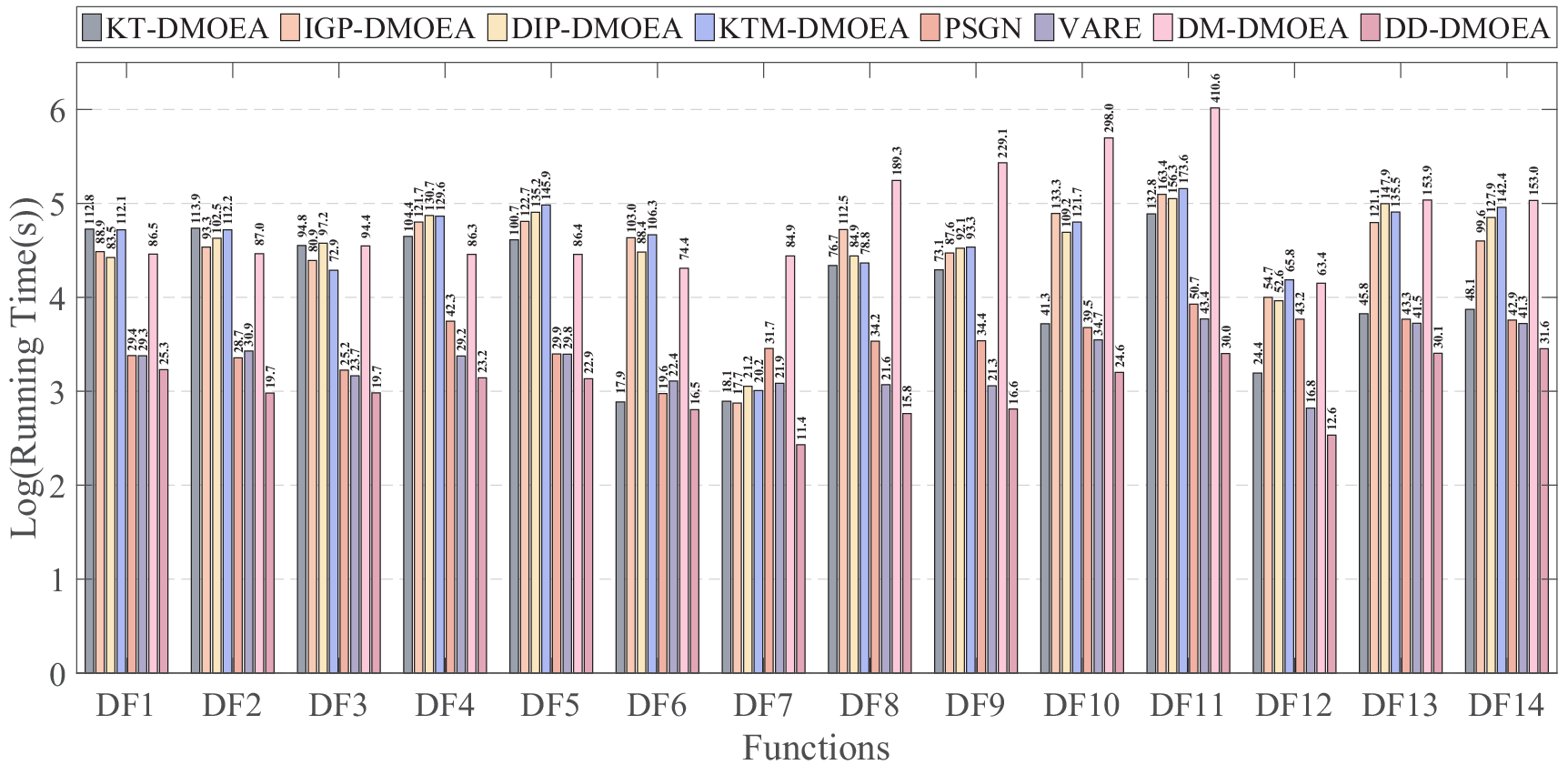}
    \caption{Average running time of all algorithms on 14 test functions on dynamic environments ($n_t=10$ and $\tau_t=10$).}
    \label{fig: rt-1-nt10-tauT10}
\end{figure*}

\begin{figure*}
    \centering
    \includegraphics[width=\linewidth]{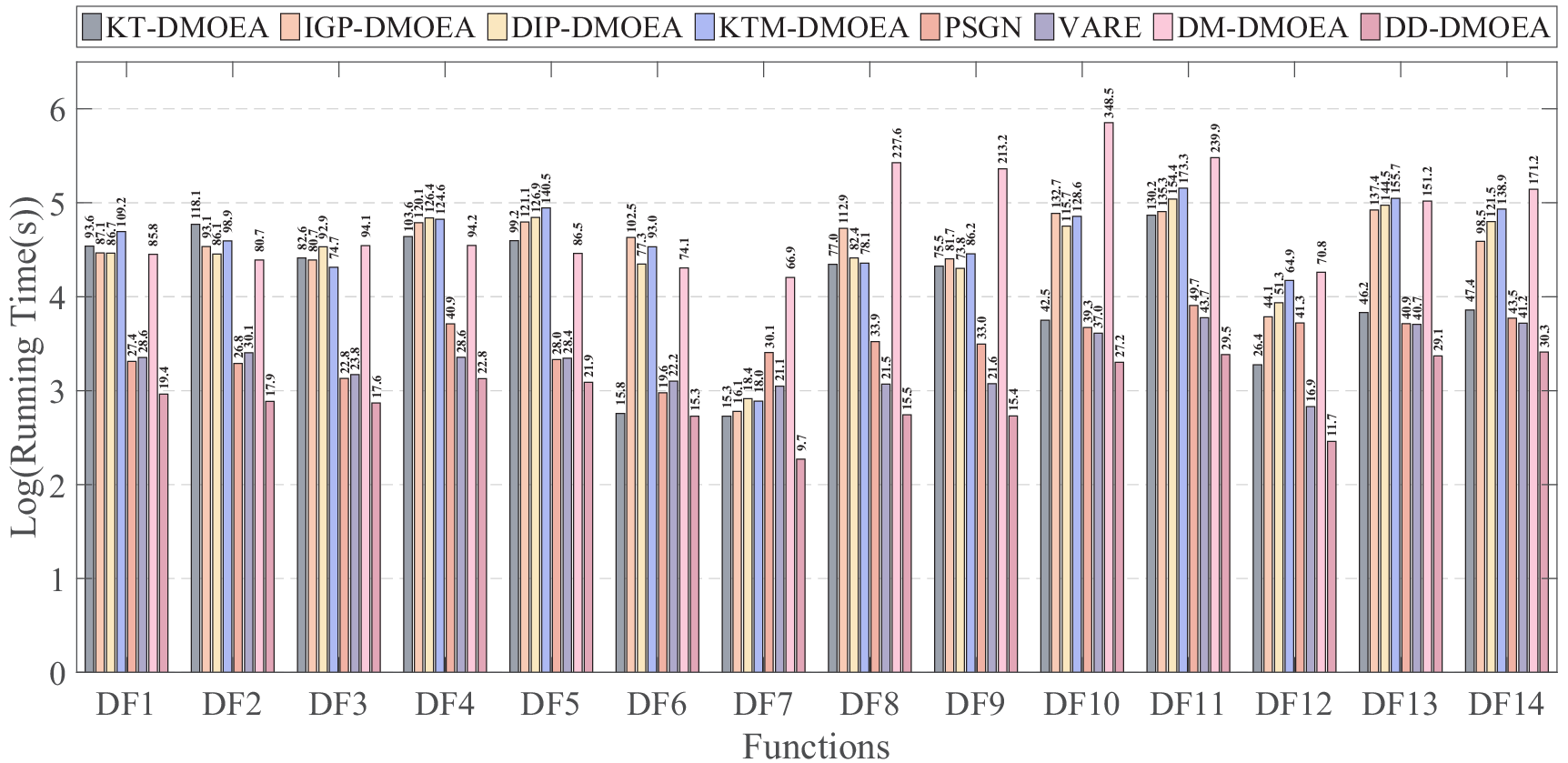}
    \caption{Average running time of all algorithms on 14 test functions on dynamic environments ($n_t=10$ and $\tau_t=5$).}
    \label{fig: rt-2-nt5-tauT10}
\end{figure*}

\begin{figure*}
    \centering
    \includegraphics[width=\linewidth]{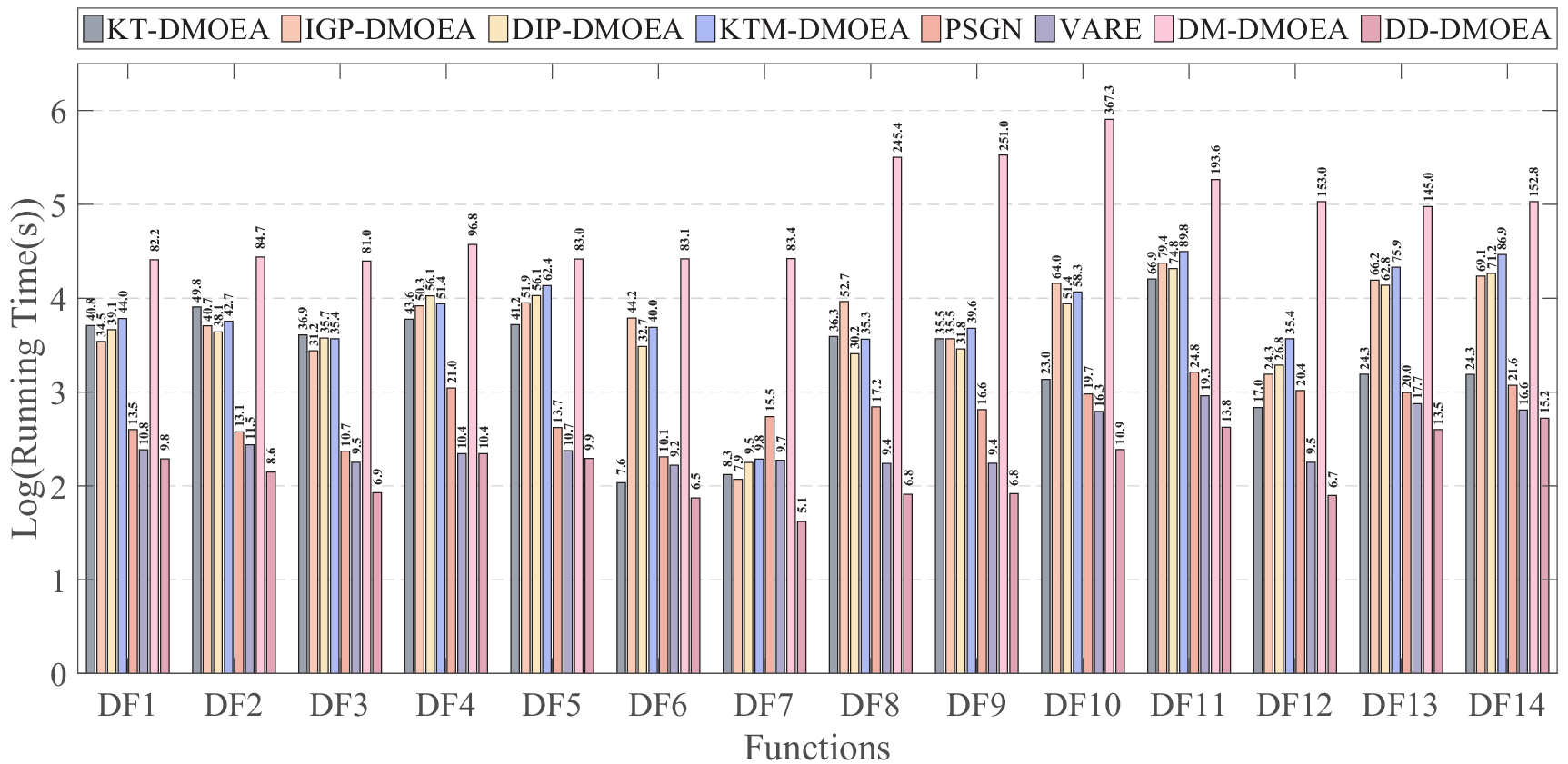}
    \caption{Average running time of all algorithms on 14 test functions on dynamic environments ($n_t=5$ and $\tau_t=10$).}
    \label{fig: rt-3-nt10-tauT5}
\end{figure*}

\begin{figure*}
    \centering
    \includegraphics[width=\linewidth]{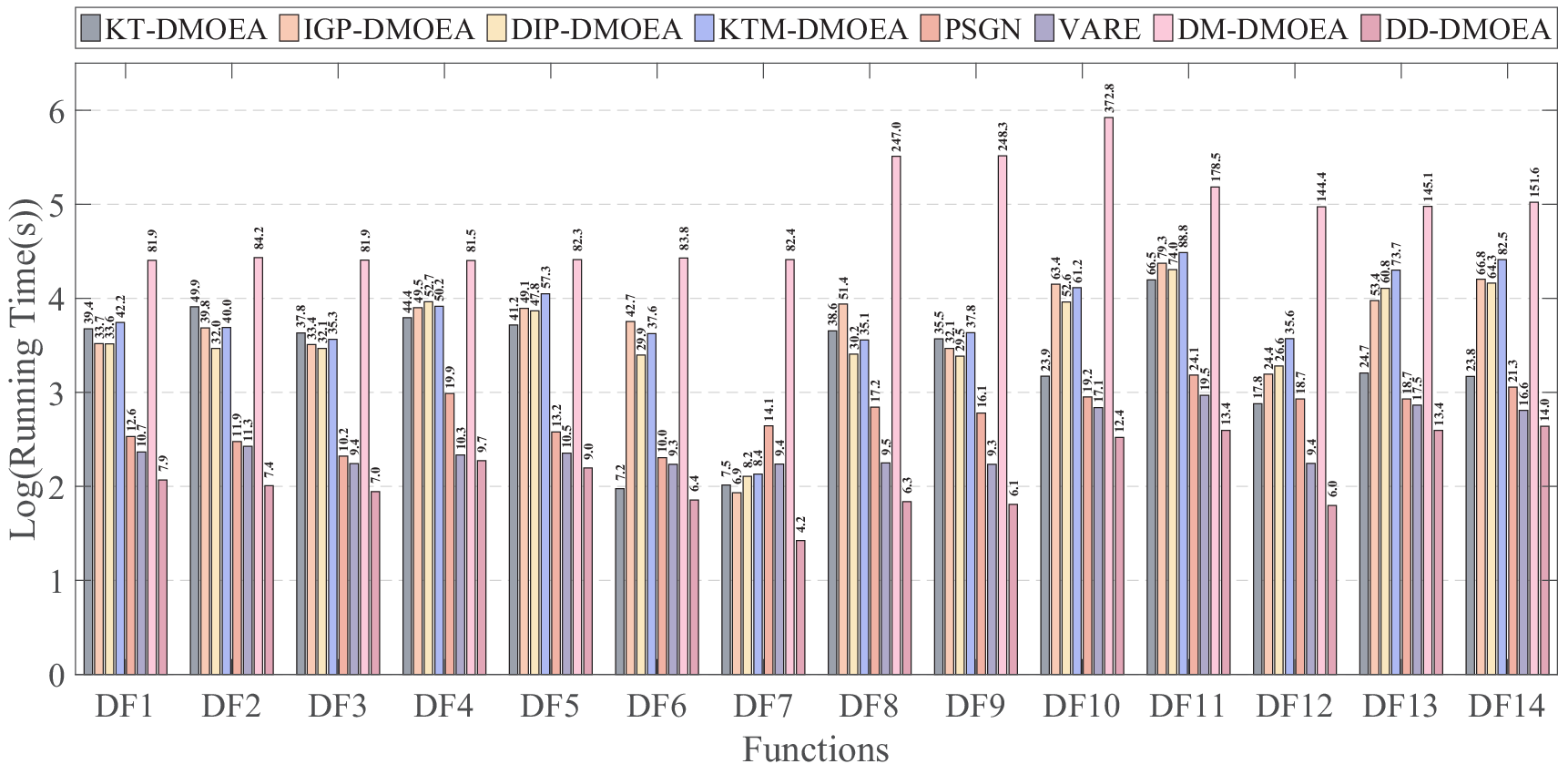}
    \caption{Average running time of all algorithms on 14 test functions on dynamic environments ($n_t=5$ and $\tau_t=5$).}
    \label{fig: rt-4-nt5-tauT5}
\end{figure*}

To evaluate the dynamic responsiveness of each algorithm, Figs. \ref{fig: rt-1-nt10-tauT10} to \ref{fig: rt-4-nt5-tauT5} systematically presents the detailed runtimes of the algorithms in 14 test functions in four dynamic environments. The experimental data show that, under the four environmental parameter settings, the average dynamic response times of IGP-DMOEA, DIP-DMOEA, KTM-DMOEA, and DM-DMOEA are relatively long, while the dynamic response speeds of the remaining algorithms are significantly faster. The DD-DMOEA proposed in this paper possesses the fastest dynamic response speed on many test functions, which is mainly attributed to the training-free diffusion mechanism adopted in this paper, thereby achieving one of the fastest dynamic responses. 

To further validate the effectiveness of the proposed method and parameter settings of DD-DMOEA, this study conducted detailed ablation studies and parameter sensitivity analyses, with specific data provided in Tables \ref{Ablation results-MIGD} to \ref{ParSen results-Std MHV} of the supplementary material. Specifically, the results of the ablation experiments and the sensitivity analysis of parameter $\alpha$ are summarized in Tables III to VI and are discussed in depth in the main text. Furthermore, to evaluate the impact of the key parameter $\psi$ (the standard deviation of the Gaussian distribution) on the predicted population, this paper set multiple distinct $\psi$ values ($\psi=[0.01, 0.1, 0.3, 0.5, 0.7, 0.9]$) for sensitivity testing—corresponding to DDM/$\psi_1$ through DDM/$\psi_6$ in Tables \ref{ParSen results-Std MIGD} and \ref{ParSen results-Std MHV}. However, given that the guidance effectiveness of DDM is highly dependent on the accuracy of the knee points predicted by AKP, DD-DMOEA adopts an uncertainty-aware diffusion guidance strategy. This strategy adaptively adjusts the variance of the Gaussian distribution, thereby enhancing the algorithm's robustness when facing complex environmental changes.

By comparing the MIGD and MHV metrics under different $\psi$ values in Tables \ref{ParSen results-Std MIGD} and \ref{ParSen results-Std MHV}, the statistical test results clearly indicate that DD-DMOEA achieved the optimal performance. A deeper analysis of the underlying reasons reveals that larger $\psi$ values (such as 0.7 and 0.9) cause the Gaussian distribution to become overly flat, triggering an ``over-smoothing" phenomenon in probability density. This blurring effect makes it difficult for the model to distinguish between high-quality solutions close to the knee points and low-quality solutions far from them, which in turn leads to inaccurate weighted averaging when calculating the target expected values, ultimately resulting in degraded algorithm performance.

Conversely, a smaller $\psi$ value facilitates the algorithm's focus on modeling the core region of the population. A sharper probability distribution generates strong guidance signals, compelling the population to converge tightly toward the high-quality knee point regions during the denoising process, thereby continuously guiding the population to evolve toward high-quality areas. Based on the comprehensive comparison results across multiple test functions, $\psi$ exhibits superior performance within the range of $[0.1, 0.5]$. Consequently, this paper sets $\psi_{\min}$ and $\psi_{\max}$ to 0.1 and 0.5, respectively. Overall, the comprehensive experimental results fully demonstrate that the core methodological framework and parameter settings adopted by DD-DMOEA exhibit significant performance advantages in dynamic multiobjective optimization scenarios, thereby fully verifying its effectiveness and robustness.
\begin{table*}[ht]
  \centering
  \caption{Mean and standard deviation of MHV obtained by the comparison algorithms in four dynamic environments.}
    \resizebox{\textwidth}{!}{
    \begin{tabular}{lccccccccc}
    \toprule
    Prob. & ($n_t,\tau_t$)   & KT-DMOEA & IGP-DMOEA & DIP-DMOEA & KTM-DMOEA & PSGN  & VARE  & DM-DMOEA & DD-DMOEA \\
    \midrule
    \multirow{4}{*}{DF1} 
    & (10, 10) & 0.5854±1.19e-02(+) & 0.6079±6.27e-03(+) & 0.6038±4.04e-03(+) & 0.6089±3.67e-03(+) & 0.4666±5.34e-03(+) & 0.5783±7.94e-03(+) & 0.6053±6.33e-03(+) & \textbf{0.6096±1.29e-03} \\
    & (5, 10) & 0.5605±1.52e-02(+) & \textbf{0.5988±7.57e-03(=)} & 0.5854±6.22e-03(+) & 0.5956±6.00e-03(+) & 0.4689±5.97e-03(+) & 0.5767±8.36e-03(+) & 0.5973±7.54e-03(=) & 0.5971±1.46e-03 \\
    & (10, 5) & 0.4706±2.47e-02(+) & 0.5366±1.07e-02(+) & 0.5390±1.20e-02(+) & 0.5457±1.10e-02(+) & 0.4359±7.94e-03(+) & 0.4469±1.42e-02(+) & 0.5454±1.04e-02(+) & \textbf{0.5547±3.89e-03} \\
    & (5, 5) & 0.4358±2.95e-02(+) & \textbf{0.5126±1.46e-02(=)} & 0.4731±1.33e-02(+) & 0.5149±8.83e-03(=) & 0.4220±1.13e-02(+) & 0.4418±1.52e-02(+) & 0.5204±1.55e-02(=) & 0.5103±3.14e-03 \\
    \midrule
    \multirow{4}{*}{DF2}     
    & (10, 10) & 0.8125±1.10e-02(+) & 0.8291±8.00e-03(+) & 0.8245±5.74e-03(+) & 0.8312±4.56e-03(+) & 0.6634±6.50e-03(+) & 0.8182±3.99e-03(+) & \textbf{0.8368±8.88e-03(=)} & 0.8357±1.14e-03 \\
    & (5, 10) & 0.8016±1.64e-02(+) & 0.8214±1.01e-02(=) & 0.7992±8.82e-03(+) & 0.8127±6.28e-03(+) & 0.6638±6.10e-03(+) & 0.8134±5.73e-03(+) & \textbf{0.8222±1.10e-02(=)} & 0.8184±2.17e-03 \\
    & (10, 5) & 0.6936±2.44e-02(+) & 0.7474±1.41e-02(+) & 0.7458±1.49e-02(+) & 0.7466±1.21e-02(+) & 0.6493±7.70e-03(+) & 0.7140±9.71e-03(+) & 0.7596±8.57e-03(+) & \textbf{0.7673±3.44e-03} \\
    & (5, 5) & 0.6777±3.12e-02(+) & 0.7231±1.60e-02(+) & 0.6791±2.25e-02(+) & 0.7053±1.56e-02(+) & 0.6437±6.61e-03(+) & 0.6999±1.64e-02(+) & 0.7242±8.90e-03(+) & \textbf{0.7255±3.22e-03} \\
    \midrule
    \multirow{4}{*}{DF3}     
    & (10, 10) & 0.3735±1.16e-02(+) & 0.4831±1.04e-02(+) & \textbf{0.5020±1.13e-02(=)} & 0.4288±1.47e-02(+) & 0.2878±1.79e-02(+) & 0.3747±1.38e-02(+) & 0.4928±1.21e-02(+) & 0.5005±6.73e-03 \\
    & (5, 10) & 0.3678±1.29e-02(+) & 0.4490±1.42e-02(=) & \textbf{0.4859±1.54e-02(-)} & 0.4226±1.30e-02(+) & 0.2786±1.92e-02(+) & 0.3714±1.80e-02(+) & 0.4820±1.31e-02(-) & 0.4450±6.19e-03 \\
    & (10, 5) & 0.2858±1.47e-02(+) & 0.3709±1.39e-02(=) & \textbf{0.4252±2.72e-02(-)} & 0.3286±2.28e-02(+) & 0.1883±1.74e-02(+) & 0.2756±2.73e-02(+) & 0.3922±7.25e-03(-) & 0.3674±1.01e-02 \\
    & (5, 5) & 0.2832±1.63e-02(+) & 0.3474±1.31e-02(=) & \textbf{0.3760±2.58e-02(-)} & 0.3192±1.70e-02(+) & 0.1832±2.68e-02(+) & 0.2706±2.04e-02(+) & 0.3586±2.21e-02(-) & 0.3406±6.22e-03 \\
    \midrule
    \multirow{4}{*}{DF4}     
    & (10, 10) & 5.9070±2.51e-02(+) & 5.9879±1.97e-02(=) & 5.9235±1.54e-02(+) & 5.9201±1.53e-02(+) & 0.6473±1.79e-02(+) & 6.0448±3.37e-02(=) & \textbf{6.2066±3.16e-02(=)} & 5.9445±3.75e-03 \\
    & (5, 10) & 5.9847±2.78e-02(+) & 6.0485±2.16e-02(+) & 6.0599±2.03e-02(+) & 6.0582±1.52e-02(+) & 0.6528±1.98e-02(+) & 5.9037±4.11e-02(+) & 6.0495±3.29e-02(+) & \textbf{6.0885±3.44e-03} \\
    & (10, 5) & 5.3980±1.13e-01(+) & 5.7953±7.90e-02(=) & 5.6846±8.83e-02(+) & 5.6371±6.50e-02(+) & 0.5641±3.31e-02(+) & 5.2534±1.47e-01(+) & \textbf{6.0120±4.72e-02(=)} & 5.7866±3.57e-02 \\
    & (5, 5) & 5.4884±1.20e-01(+) & 5.7811±1.04e-01(+) & 5.7949±8.26e-02(+) & 5.7395±7.01e-02(+) & 0.5481±3.91e-02(+) & 5.1335±1.14e-01(+) & 5.8056±7.61e-02(+) & \textbf{5.8803±3.86e-02} \\
    \midrule
    \multirow{4}{*}{DF5}     
    & (10, 10) & 0.6509±1.66e-02(+) & 0.6802±3.33e-03(=) & 0.6806±2.23e-03(=) & \textbf{0.6834±2.09e-03(=)} & 0.4619±2.03e-02(+) & 0.6428±8.55e-03(+) & 0.6797±4.81e-03(=) & 0.6786±5.27e-04 \\
     & (5, 10) & 0.6532±9.93e-03(+) & 0.6790±4.01e-03(=) & 0.6776±5.59e-03(=) & \textbf{0.6813±2.75e-03(=)} & 0.4630±1.93e-02(+) & 0.6372±7.63e-03(+) & 0.6794±5.37e-03(=) & 0.6761±7.69e-04 \\
    & (10, 5) & 0.4998±2.92e-02(+) & 0.6352±1.27e-02(=) & 0.6430±6.47e-03(=) & \textbf{0.6456±8.41e-03(=)} & 0.4168±1.69e-02(+) & 0.4738±3.04e-02(+) & 0.6415±7.97e-03(=) & 0.6341±4.15e-03 \\
    & (5, 5) & 0.4983±2.93e-02(+) & 0.6172±1.24e-02(=) & 0.6182±1.35e-02(=) & \textbf{0.6324±1.41e-02(=)} & 0.3754±2.39e-02(+) & 0.4612±3.08e-02(+) & 0.6200±1.55e-02(=) & 0.6169±2.41e-03 \\
    \midrule
    \multirow{4}{*}{DF6}     
    & (10, 10) & 0.5228±6.79e-02(+) & 0.5618±7.09e-02(+) & 0.5846±1.42e-02(+) & 0.5692±5.34e-02(+) & 0.1475±7.70e-02(+) & 0.4520±4.45e-02(+) & 0.5901±8.60e-03(+) & \textbf{0.6009±4.62e-03} \\
    & (5, 10) & 0.4036±6.61e-02(+) & 0.5922±2.30e-02(+) & 0.5696±2.69e-02(+) & 0.5609±3.49e-02(+) & 0.1154±3.33e-02(+) & 0.4333±4.56e-02(+) & 0.5600±3.55e-02(+) & \textbf{0.5955±3.45e-03} \\
    & (10, 5) & 0.3583±1.27e-01(+) & 0.5180±4.93e-02(+) & 0.5216±2.54e-02(+) & 0.4780±6.04e-02(+) & 0.0547±4.65e-02(+) & 0.3314±4.72e-02(+) & 0.5312±1.61e-02(+) & \textbf{0.5456±9.25e-03} \\
    & (5, 5) & 0.2086±6.88e-02(+) & \textbf{0.5119±5.83e-02(=)} & 0.4397±5.08e-02(+) & 0.4049±8.11e-02(+) & 0.0536±3.40e-02(+) & 0.3033±4.84e-02(+) & 0.4828±4.19e-02(+) & 0.4917±1.10e-02 \\
    \midrule
    \multirow{4}{*}{DF7}     
    & (10, 10) & 1.9071±2.37e-02(+) & 1.9283±1.22e-02(+) & 2.0117±3.96e-02(=) & 1.8907±3.56e-01(+) & 0.3890±7.85e-03(+) & 1.6571±5.73e-02(+) & \textbf{2.0942±3.07e-02(-)} & 1.9829±7.64e-02 \\
    & (5, 10) & 1.6918±2.68e-02(+) & 1.7003±1.78e-02(+) & 1.8252±8.43e-02(=) & 1.7924±3.44e-01(=) & 0.3798±1.13e-02(+) & 1.5371±4.05e-02(+) & \textbf{1.8887±6.99e-02(-)} & 1.7918±4.46e-02 \\
    & (10, 5) & 1.6891±6.25e-02(=) & 1.8348±1.96e-02(-) & 1.9235±4.08e-02(-) & 1.8040±3.49e-01(-) & 0.3770±1.17e-02(+) & 1.5070±5.64e-02(+) & \textbf{1.8926±1.04e-01(-)} & 1.6073±1.89e-02 \\
    & (5, 5) & 1.5088±4.35e-02(+) & 1.6357±5.49e-02(+) & \textbf{1.7252±8.72e-02(=)} & 1.6915±3.04e-01(+) & 0.3700±1.70e-02(+) & 1.4161±4.11e-02(+) & 1.6952±6.80e-02(+) & 1.7023±2.19e-02 \\
    \midrule
    \multirow{4}{*}{DF8}     
    & (10, 10) & 0.6674±9.88e-03(+) & 0.7013±6.21e-03(+) & 0.7102±4.95e-03(+) & 0.6758±1.03e-02(+) & 0.5862±7.20e-03(+) & 0.6600±7.21e-03(+) & 0.7067±3.55e-03(+) & \textbf{0.7104±2.16e-03} \\
    & (5, 10) & 0.6709±1.32e-02(+) & 0.6969±1.36e-02(+) & 0.7087±4.55e-03(+) & 0.6767±1.33e-02(+) & 0.5901±4.46e-03(+) & 0.6612±1.13e-02(+) & 0.7087±3.57e-03(+) & \textbf{0.7150±2.36e-03} \\
    & (10, 5) & 0.6066±1.98e-02(+) & \textbf{0.6886±8.62e-03(=)} & 0.6721±9.78e-03(+) & 0.5983±1.40e-02(+) & 0.5601±5.61e-03(+) & 0.5934±1.16e-02(+) & 0.6690±8.98e-03(+) & 0.6871±4.27e-03 \\
    & (5, 5) & 0.5991±1.80e-02(+) & 0.6814±8.68e-03(+) & 0.6803±6.71e-03(+) & 0.5916±1.32e-02(+) & 0.5672±8.35e-03(+) & 0.5882±1.08e-02(+) & 0.6684±7.35e-03(+) & \textbf{0.6827±3.59e-03} \\
    \midrule
    \multirow{4}{*}{DF9}     
    & (10, 10) & 0.4933±1.99e-02(+) & 0.5462±1.88e-02(=) & \textbf{0.5615±1.46e-02(=)} & 0.5371±1.97e-02(+) & 0.4232±1.38e-02(+) & 0.4420±1.64e-02(+) & 0.5092±2.29e-02(+) & 0.5454±1.11e-02 \\
    & (5, 10) & 0.4708±2.79e-02(+) & 0.5118±3.25e-02(+) & 0.4775±1.41e-02(+) & 0.5270±1.28e-02(+) & 0.4055±1.03e-02(+) & 0.4451±1.90e-02(+) & 0.4879±2.96e-02(+) & \textbf{0.5341±6.52e-03} \\
    & (10, 5) & 0.3067±3.56e-02(+) & 0.3979±2.16e-02(+) & 0.4207±2.08e-02(+) & 0.4384±2.20e-02(+) & 0.3119±1.87e-02(+) & 0.3140±2.19e-02(+) & 0.4424±3.75e-02(+) & \textbf{0.4657±5.37e-03} \\
    & (5, 5) & 0.2963±3.02e-02(+) & 0.3374±3.03e-02(+) & 0.3776±2.73e-02(+) & 0.3976±2.84e-02(+) & 0.2689±2.63e-02(+) & 0.3128±2.22e-02(+) & 0.3823±4.54e-02(+) & \textbf{0.3986±1.06e-02} \\
    \midrule
    \multirow{4}{*}{DF10}    
    & (10, 10) & 0.6382±1.45e-02(+) & 0.6281±1.33e-02(+) & 0.7351±2.93e-02(+) & 0.6878±1.29e-02(+) & 0.4570±8.53e-03(+) & 0.6860±1.25e-02(+) & 0.7314±1.69e-02(+) & \textbf{0.7598±1.07e-02} \\
    & (5, 10) & 0.7469±1.26e-02(+) & 0.6280±1.37e-01(+) & 0.8424±2.24e-02(+) & 0.8260±1.05e-02(+) & 0.5610±8.25e-03(+) & 0.8257±1.21e-02(+) & 0.8735±1.14e-02(+) & \textbf{0.8841±9.22e-03} \\
    & (10, 5) & 0.5484±1.62e-02(+) & 0.5846±3.31e-02(+) & 0.7139±3.24e-02(+) & 0.6149±1.79e-02(+) & 0.4286±1.75e-02(+) & 0.5836±1.60e-02(+) & 0.7050±1.54e-02(+) & \textbf{0.7491±1.77e-02} \\
    & (5, 5) & 0.6693±1.32e-02(+) & 0.5577±1.54e-01(+) & 0.8000±2.30e-02(+) & 0.7584±1.11e-02(+) & 0.5426±2.16e-02(+) & 0.7206±1.70e-02(+) & 0.8189±1.52e-02(+) & \textbf{0.8603±8.67e-03} \\
    \midrule
    \multirow{4}{*}{DF11}     
    & (10, 10) & 0.8685±6.46e-03(+) & 0.8811±8.47e-03(+) & 1.0328±8.28e-03(+) & 1.0020±8.21e-03(+) & 0.1997±5.59e-03(+) & 0.8823±7.06e-03(+) & 0.8273±6.82e-03(+) & \textbf{1.0339±4.49e-03} \\
    & (5, 10) & 0.8621±9.89e-03(+) & 0.8673±1.13e-02(+) & 0.9997±6.59e-03(+) & 0.9665±6.69e-03(+) & 0.1960±3.70e-03(+) & 0.8753±7.15e-03(+) & 0.8223±8.99e-03(+) & \textbf{1.0018±4.61e-03} \\
    & (10, 5) & 0.7216±1.54e-02(+) & 0.7598±1.88e-02(+) & 0.9561±1.65e-02(+) & 0.9307±1.43e-02(+) & 0.2041±7.54e-03(+) & 0.7300±2.01e-02(+) & 0.7802±7.91e-03(+) & \textbf{0.9867±8.24e-03} \\
    & (5, 5) & 0.7142±1.72e-02(+) & 0.7423±1.56e-02(+) & 0.9214±1.58e-02(+) & 0.8990±1.64e-02(+) & 0.1863±6.16e-03(+) & 0.7434±1.98e-02(+) & 0.7472±4.26e-03(+) & \textbf{0.9640±7.56e-03} \\
    \midrule
    \multirow{4}{*}{DF12}     
    & (10, 10) & 1.2669±4.03e-02(+) & 1.2260±1.18e-01(+) & 1.3040±2.34e-09(+) & 1.3040±3.28e-07(+) & 0.3165±1.03e-02(+) & 1.3170±3.40e-10(=) & \textbf{1.3170±0.00e+00(=)} & 1.3040±2.34e-16 \\
    & (5, 10) & 1.1296±5.23e-02(+) & 1.2611±7.59e-02(+) & 1.2895±7.40e-10(+) & 1.2895±1.69e-10(+) & 0.2730±1.27e-02(+) & 1.2895±4.83e-10(+) & 1.2895±0.00e+00(+) & \textbf{1.2895±2.34e-16} \\
    & (10, 5) & 1.2649±3.19e-02(+) & 1.2520±1.07e-01(+) & 1.3040±6.49e-05(+) & 1.3011±6.98e-03(+) & 0.2689±1.62e-02(+) & 1.3170±2.41e-05(=) & \textbf{1.3170±6.05e-08(=)} & 1.3040±5.78e-07 \\
    & (5, 5) & 1.1430±7.19e-02(+) & 1.1443±2.55e-01(+) & 1.2895±2.06e-05(+) & 1.2885±2.11e-03(+) & 0.2142±2.43e-02(+) & 1.2895±1.17e-05(+) & 1.2895±2.12e-07(+) & \textbf{1.2895±4.70e-07} \\
    \midrule
    \multirow{4}{*}{DF13}     
    & (10, 10) & 3.1951±5.58e-02(+) & \textbf{3.3263±3.59e-02(=)} & 3.2527±2.37e-02(+) & 3.3123±2.67e-02(=) & 0.5421±1.05e-02(+) & 3.1780±2.20e-02(+) & 3.2731±2.34e-02(+) & 3.2916±8.63e-03 \\
    & (5, 10) & 3.2202±4.63e-02(+) & 3.3383±2.96e-02(=) & 3.3079±2.64e-02(+) & \textbf{3.3884±2.24e-02(=)} & 0.5400±1.27e-02(+) & 3.2525±3.47e-02(+) & 3.3593±1.91e-02(=) & 3.3339±1.18e-02 \\
    & (10, 5) & 2.5506±1.03e-01(+) & 3.1622±4.32e-02(=) & 3.0438±5.28e-02(+) & \textbf{3.1861±3.68e-02(=)} & 0.4658±2.12e-02(+) & 2.7168±4.69e-02(+) & 3.0243±5.65e-02(+) & 3.0894±2.19e-02 \\
    & (5, 5) & 2.6388±9.26e-02(+) & 3.0854±5.62e-02(=) & 2.9854±7.22e-02(+) & \textbf{3.2214±2.87e-02(-)} & 0.4945±1.68e-02(+) & 2.7003±5.91e-02(+) & 3.0870±2.06e-02(=) & 3.0338±2.47e-02 \\
    \midrule
    \multirow{4}{*}{DF14}     
    & (10, 10) & 0.5420±8.72e-03(+) & 0.5640±4.30e-03(+) & 0.5648±2.15e-03(+) & 0.5620±2.77e-03(+) & 0.0439±2.92e-02(+) & 0.5161±4.16e-03(+) & 0.5331±2.80e-03(+) & \textbf{0.5681±7.97e-04} \\
    & (5, 10) & 0.5213±7.99e-03(+) & 0.5415±3.51e-03(+) & 0.5413±1.86e-03(+) & 0.5401±2.86e-03(+) & 0.0379±2.39e-02(+) & 0.5293±5.49e-03(+) & \textbf{0.5464±1.74e-03(=)} & 0.5444±7.01e-04 \\
    & (10, 5) & 0.3976±2.20e-02(+) & 0.5233±7.83e-03(+) & 0.5403±6.51e-03(+) & 0.5373±5.93e-03(+) & 0.0310±2.24e-02(+) & 0.4213±1.89e-02(+) & 0.5075±3.70e-03(+) & \textbf{0.5409±1.59e-03} \\
    & (5, 5) & 0.3761±2.92e-02(+) & 0.4982±1.05e-02(+) & 0.4986±1.02e-02(+) & 0.5085±8.68e-03(=) & 0.0310±2.19e-02(+) & 0.4140±1.59e-02(+) & \textbf{0.5151±7.14e-03(=)} & 0.5059±5.32e-03 \\
    \midrule
    \multicolumn{2}{c}{+/=/-} & 55/1/0 & 36/19/1 & 43/9/4 & 44/10/2 & 56/0/0 & 53/3/0 & 34/16/6 & - \\
    \bottomrule
    \end{tabular}%
    }
  \label{tab: MHV of comparison algorithms}%
\end{table*}%

\begin{table*}[h]
    \centering
    \scriptsize
    \caption{Mean and standard deviation of MIGD values of all test functions for different strategies on four dynamic environments.}
    \label{Ablation results-MIGD}
\begin{tabular}{lccccc}
    \toprule
    Prob. & $(n_t,\tau_t)$  & DD-DMOEA/$\text{v}_1$ & DD-DMOEA$/\text{v}_2$ & DD-DMOEA$/\text{v}_3$ & DD-DMOEA \\
    \midrule
    \multirow{4}{*}{DF1}
    & (10, 10) & 0.0251±2.92e-03(+)  & 0.0394±4.01e-03(+)  & 0.0312±2.17e-03(+)  & \textbf{0.0235±2.09e-03}  \\
    & (5, 10)  & 0.0374±3.86e-03(+)  & 0.0472±5.75e-03(+)  & 0.0404±3.76e-03(+)  & \textbf{0.0356±3.91e-03}  \\
    & (10, 5)  & \textbf{0.0571±5.50e-03(=)}  & 0.1338±1.39e-02(+)  & 0.0910±4.30e-03(+)  & 0.0576±5.87e-03  \\
    & (5, 5)   & 0.0934±7.35e-03(+)  & 0.1372±1.30e-02(+)  & 0.1130±1.14e-02(+)  & \textbf{0.0872±9.67e-03}  \\
    \midrule
    \multirow{4}{*}{DF2}      
    & (10, 10) & 0.0264±4.45e-03(+)  & 0.0388±6.20e-03(+)  & 0.0253±4.03e-03(+)  & \textbf{0.0252±4.58e-03}  \\
    & (5, 10)  & 0.0346±4.99e-03(+)  & 0.0392±2.89e-03(+)  & \textbf{0.0298±2.91e-03(-)}  & 0.0327±5.70e-03  \\
    & (10, 5)  & 0.0675±6.53e-03(+)  & 0.1185±1.25e-02(+)  & 0.0752±6.16e-03(+)  & \textbf{0.0636±2.82e-03}  \\
\bottomrule
    \end{tabular}
\end{table*}

\begin{table*}[ht]
    \centering
    \scriptsize
    \caption{Mean and standard deviation of MHV values of all test functions for different strategies on four dynamic environments.}
    \label{Ablation results-MHV}
    \begin{tabular}{lccccc}
    \toprule
    Prob. & $(n_t,\tau_t)$  & DD-DMOEA/$\text{v}_1$ & DD-DMOEA$/\text{v}_2$ & DD-DMOEA$/\text{v}_3$ & DD-DMOEA\\
    \midrule
    \multirow{4}{*}{DF1}
    & (10, 10) & 0.6057±5.12e-03(+) & 0.5800±6.93e-03(+) & 0.5945±3.93e-03(+) & \textbf{0.6081±3.66e-03} \\
    & (5, 10)  & 0.5904±7.15e-03(+) & 0.5722±9.67e-03(+) & 0.5829±8.10e-03(+) & \textbf{0.5936±8.81e-03} \\
    & (10, 5)  & \textbf{0.5501±8.02e-03(=)} & 0.4359±1.69e-02(+) & 0.4934±5.33e-03(+) & 0.5486±8.66e-03 \\
    & (5, 5)   & 0.4953±1.25e-02(+) & 0.4285±1.52e-02(+) & 0.4607±1.75e-02(+) & \textbf{0.5040±1.43e-02} \\
    \midrule
    \multirow{4}{*}{DF2}
    & (10, 10) & 0.8302±6.77e-03(+) & 0.8067±1.08e-02(+) & \textbf{0.8343±4.35e-03(=)} & 0.8316±7.90e-03 \\
    & (5, 10)  & 0.8128±1.02e-02(+) & 0.8070±5.20e-03(+) & \textbf{0.8233±5.19e-03(=)} & 0.8159±1.08e-02 \\
    & (10, 5)  & 0.7579±1.30e-02(+) & 0.6748±1.93e-02(+) & 0.7421±1.07e-02(+) & \textbf{0.7628±6.73e-03} \\
\bottomrule
    \end{tabular}
\end{table*}

\begin{table*}
    \centering
    \caption{Mean and standard deviations of MIGD values obtained from all test functions in four dynamic environments by different time steps $K$ and noise scheduling $\alpha$}
    \label{ParSen results-MIGD}
    \resizebox{\textwidth}{!}{
    \begin{tabular}{lcccccccc}
    \toprule
    Prob. & $(n_t,\tau_t)$ & DDM/L($\alpha$)-1 & DDM/$K_1$ & DDM/$K_2$ & DDM/$K_3$ & DDM/$K_4$ & DDM/$K_5$ & DD-DMOEA \\
    \midrule
    \multirow{4}{*}{DF1}
    & (10, 10) & 0.0257±2.33e-03(+)  & 0.0251±2.01e-03(+)  & 0.0245±2.65e-03(+)  & 0.0249±2.28e-03(+)  & 0.0241±2.02e-03(+)  & 0.0254±2.75e-03(+)  & \textbf{0.0235±2.09e-03}  \\
    & (5, 10)  & 0.0395±4.38e-03(+)  & 0.0419±4.53e-03(+)  & 0.0362±3.26e-03(+)  & 0.0363±3.55e-03(+)  & 0.0362±2.91e-03(+)  & 0.0369±5.08e-03(+)  & \textbf{0.0356±3.91e-03}  \\
    & (10, 5)  & 0.0662±4.51e-03(+)  & 0.0619±6.57e-03(+)  & 0.0576±4.24e-03(=)  & 0.0598±6.94e-03(+)  & 0.0580±5.04e-03(+)  & 0.0603±3.99e-03(+)  & \textbf{0.0576±5.87e-03}  \\
    & (5, 5)   & 0.0872±7.04e-03(=)  & 0.0944±6.28e-03(+)  & 0.0868±7.26e-03(=)  & 0.0888±8.88e-03(+)  & 0.0846±7.99e-03(=)  & \textbf{0.0812±6.30e-03(-)}  & 0.0872±9.67e-03  \\
    \midrule
    \multirow{4}{*}{DF2}      
    & (10, 10) & 0.0267±3.26e-03(+)  & 0.0263±3.41e-03(+)  & 0.0256±2.87e-03(+)  & 0.0252±3.09e-03(=)  & 0.0249±4.55e-03(=)  & 0.0259±4.54e-03(+)  & \textbf{0.0252±4.58e-03}  \\
    & (5, 10)  & 0.0375±4.59e-03(+)  & 0.0355±4.04e-03(+)  & 0.0351±4.97e-03(+)  & 0.0341±4.30e-03(+)  & 0.0351±4.92e-03(+)  & 0.0360±4.87e-03(+)  & \textbf{0.0327±5.70e-03}  \\
    & (10, 5)  & 0.0664±7.54e-03(+)  & 0.0696±8.46e-03(+)  & 0.0663±7.19e-03(+)  & 0.0677±8.03e-03(+)  & 0.0639±6.37e-03(+)  & \textbf{0.0610±6.72e-03(=)}  & 0.0636±2.82e-03  \\
\bottomrule
    \end{tabular}
    }
\end{table*}

\begin{table*}[htb]
  \centering
  \caption{Mean and standard deviations of MHV values obtained from all test functions in four dynamic environments by different time steps $K$ and noise scheduling $\alpha$}
    \label{ParSen results-MHV}
    \resizebox{\textwidth}{!}{
    \begin{tabular}{lcccccccc}
    \toprule
    Prob. & $(n_t,\tau_t)$ & DDM/L($\alpha$)-1 & DDM/$K_1$ & DDM/$K_2$ & DDM/$K_3$ & DDM/$K_4$ & DDM/$K_5$ & DD-DMOEA \\
    \midrule
    \multirow{4}{*}{DF1}
    & (10, 10) & 0.6035±4.29e-03(+) & 0.6055±3.76e-03(+) & 0.6069±4.15e-03(+) & 0.6057±3.84e-03(+) & 0.6069±3.89e-03(+) & 0.6044±4.67e-03(+) & \textbf{0.6081±3.61e-03} \\
    & (5, 10) & 0.5849±7.85e-03(+) & 0.5817±8.11e-03(+) & 0.5921±6.73e-03(+) & 0.5926±6.42e-03(+) & 0.5925±5.09e-03(+) & 0.5919±8.78e-03(+) & \textbf{0.5936±8.21e-03} \\
    & (10, 5) & 0.5345±5.43e-03(+) & 0.5413±1.09e-02(+) & \textbf{0.5512±5.77e-03(=)} & 0.5462±9.73e-03(+) & 0.5488±7.76e-03(=) & 0.5449±6.02e-03(+) & 0.5484±8.81e-03 \\
    & (5, 5) & 0.5022±1.16e-02(+) & 0.4921±9.55e-03(+) & 0.5046±1.07e-02(=) & 0.5046±1.17e-02(=) & 0.5104±1.23e-02(=) & \textbf{0.5160±1.00e-02(=)} & 0.5040±8.76e-03 \\
    \midrule
    \multirow{4}{*}{DF2} 
    & (10, 10) & 0.8292±6.01e-03(+) & 0.8306±5.78e-03(+) & \textbf{0.8317±5.83e-03(=)} & 0.8313±5.19e-03(+) & 0.8309±7.81e-03(+) & 0.8307±7.20e-03(+) & 0.8316±7.79e-03 \\
    & (5, 10) & 0.8080±9.36e-03(+) & 0.8115±8.34e-03(+) & 0.8109±9.33e-03(+) & 0.8124±9.09e-03(+) & 0.8118±9.85e-03(+) & 0.8102±9.12e-03(+) & \textbf{0.8159±1.40e-02} \\
    & (10, 5) & 0.7590±1.30e-02(+) & 0.7572±1.39e-02(+) & 0.7622±1.31e-02(+) & 0.7562±1.31e-02(+) & 0.7623±1.12e-02(+) & \textbf{0.7678±1.12e-02(=)} & 0.7628±1.50e-02 \\
\bottomrule
    \end{tabular}%
    }
\end{table*}%

\begin{table}[htbp]
  \centering
  \caption{Mean and standard deviations of MIGD values obtained from all test functions in four dynamic environments by the standard deviation $\psi$ of different Gaussian distributions.}
  \label{ParSen results-Std MIGD}
  \resizebox{\textwidth}{!}{
    \begin{tabular}{lccccccccc}
    \toprule
    Prob. & $(n_t,\tau_t)$ & DDM/$\psi_1$ & DDM/$\psi_2$& DDM/$\psi_3$ & DDM/$\psi_4$ & DDM/$\psi_5$ & DDM/$\psi_6$ & DD-DMOEA \\
    \midrule
    \multirow{4}{*}{DF1}
    & (10, 10) & 0.0267±4.13e-03(+) & 0.0247±3.89e-03(+) & \textbf{0.0227±4.13e-03(=)} & 0.0286±4.26e-03(+) & 0.0241±3.11e-03(+) & 0.0247±3.42e-03(+) & 0.0230±2.36e-03 \\
    & (5, 10) & 0.0364±4.53e-03(+) & 0.0359±5.33e-03(+) & \textbf{0.0321±4.24e-03(-)} & 0.0403±6.02e-03(+) & 0.0389±4.87e-03(+) & 0.0399±5.97e-03(+) & 0.0349±3.34e-03 \\
    & (10, 5) & 0.0786±1.26e-02(+) & 0.0727±1.07e-02(+) & 0.0644±9.43e-03(+) & 0.0826±6.54e-03(+) & 0.0669±7.74e-03(+) & 0.0646±5.91e-03(+) & \textbf{0.0566±6.05e-03} \\
    & (5, 5) & 0.1047±1.21e-02(+) & 0.1035±1.02e-02(+) & 0.0913±1.15e-02(+) & 0.1131±1.57e-02(+) & 0.1041±8.69e-03(+) & 0.1042±1.18e-02(+) & \textbf{0.0842±6.18e-03} \\
    \midrule
    \multirow{4}{*}{DF2}    
    & (10, 10) & 0.0380±6.97e-03(+) & 0.0340±6.48e-03(+) & 0.0291±5.23e-03(+) & 0.0359±4.59e-03(+) & 0.0336±4.80e-03(+) & 0.0333±3.44e-03(+) & \textbf{0.0243±4.27e-03} \\
    & (5, 10) & 0.0391±4.75e-03(+) & 0.0385±5.81e-03(+) & 0.0363±6.82e-03(+) & 0.0452±6.32e-03(+) & 0.0457±5.42e-03(+) & 0.0470±5.96e-03(+) & \textbf{0.0317±7.79e-03} \\
    & (10, 5) & 0.0856±1.19e-02(+) & 0.0823±1.04e-02(+) & 0.0724±8.96e-03(+) & 0.0865±8.90e-03(+) & 0.0825±1.15e-02(+) & 0.0791±9.44e-03(+) & \textbf{0.0623±8.92e-03} \\
\bottomrule
    \end{tabular}%
}
\end{table}%

\begin{table}[htbp]
  \centering
  \caption{Mean and standard deviations of MHV values obtained from all test functions in four dynamic environments by the standard deviation $\psi$ of different Gaussian distributions.}
  \label{ParSen results-Std MHV}
  \resizebox{\textwidth}{!}{
    \begin{tabular}{lccccccccc}
    \toprule
    Prob. & $(n_t,\tau_t)$ & DDM/$\psi_1$ & DDM/$\psi_2$& DDM/$\psi_3$ & DDM/$\psi_4$ & DDM/$\psi_5$ & DDM/$\psi_5$ & DD-DMOEA \\
    \midrule
    \multirow{4}{*}{DF1}
    & (10, 10) & 0.5985±7.46e-03(+) & 0.5999±5.74e-03(+) & 0.6041±6.85e-03(+) & 0.6021±6.81e-03(+) & 0.6057±5.21e-03(+) & 0.6047±5.41e-03(+) & \textbf{0.6088±3.61e-03} \\
    & (5, 10) & 0.5841±1.04e-02(+) & 0.5848±8.61e-03(+) & 0.5917±7.71e-03(+) & 0.5889±7.37e-03(+) & 0.5862±8.81e-03(+) & 0.5864±1.02e-02(+) & \textbf{0.5963±8.21e-03} \\
    & (10, 5) & 0.5149±1.01e-02(+) & 0.5271±1.65e-02(+) & 0.5407±1.35e-02(+) & 0.5332±1.84e-02(+) & 0.5428±1.17e-02(+) & 0.5436±7.73e-03(+) & \textbf{0.5511±8.81e-03} \\
    & (5, 5) & 0.4886±1.89e-02(+) & 0.4857±1.36e-02(+) & 0.5016±1.56e-02(+) & 0.4940±1.46e-02(+) & 0.4907±1.29e-02(+) & 0.4945±1.67e-02(+) & \textbf{0.5075±8.76e-03} \\
    \midrule
    \multirow{4}{*}{DF2}    
    & (10, 10) & 0.8263±7.41e-03(+) & 0.8251±8.75e-03(+) & 0.8301±8.09e-03(+) & 0.8284±9.65e-03(+) & 0.8273±8.74e-03(+) & 0.8271±5.77e-03(+) & \textbf{0.8329±7.79e-03} \\
    & (5, 10) & 0.8094±1.15e-02(+) & 0.8109±1.14e-02(+) & 0.8145±1.11e-02(+) & 0.8145±9.36e-03(+) & 0.8065±1.01e-02(+) & 0.8035±1.25e-02(+) & \textbf{0.8178±1.40e-02} \\
    & (10, 5) & 0.7515±1.32e-02(+) & 0.7476±1.38e-02(+) & 0.7584±1.81e-02(+) & 0.7542±1.76e-02(+) & 0.7587±1.92e-02(+) & 0.7563±1.39e-02(+) & \textbf{0.7653±1.50e-02} \\
\bottomrule
    \end{tabular}%
    }
\end{table}%

\bibliography{mybibtex}